\documentclass[10pt]{article} % For LaTeX2e
\usepackage[preprint]{tmlr}
\usepackage{hyperref}
\usepackage{url}
\usepackage{amsfonts}       % blackboard math symbols
\usepackage{amsmath}
\usepackage{bm}
\usepackage{dsfont}
\usepackage{booktabs}
\usepackage{algorithm}
\usepackage{algorithmic}
\usepackage{graphicx}
\usepackage{caption}% http://ctan.org/pkg/caption

\ifdefined\N                                                                % N, naturals
\renewcommand{\N}{\mathds{N}}                                                % N defined by "siunitx" (which we use), for "NEWTON"
\else
  \newcommand{\N}{\mathds{N}}
\fi
\newcommand{\R}{\mathds{R}}                                                 % R, reals
\ifdefined\C
  \renewcommand{\C}{\mathds{C}}                                             % C, complex
\else
  \newcommand{\C}{\mathds{C}}
\fi
\def\argmin{\mathop{\sf arg\,min}}                                          % argmin
\newcommand{\order}{\mathcal{O}}                                            % O, order
\newcommand{\xv}{\mathbf{x}}													% vector x (bold)
\renewcommand{\P}{\mathds{P}}                                               % P, probability
\newcommand{\E}{\mathds{E}}                                                 % E, expectation
\renewcommand{\xi}[1][i]{\mathbf{x}^{(#1)}}                                          % x^i, i-th observed value of x
\newcommand{\lambdab}{\boldsymbol{\lambda}}									% input 
\title{Improving Accuracy of Interpretability Measures in Hyperparameter Optimization via Bayesian Algorithm Execution}

\author{\name Julia Moosbauer \email Julia.Moosbauer@stat.uni-muenchen.de \\
      \addr Institute of Statistics, Munich Center for Machine Learning (MCML) \\
      Ludwig-Maximilians-Universität München
      \UND
      \name Giuseppe Casalicchio \email Giuseppe.Casalicchio@stat.uni-muenchen.de \\
      \addr Institute of Statistics, Munich Center for Machine Learning (MCML) \\
        Ludwig-Maximilians-Universität München
      \UND
      \name Marius Lindauer \email lindauer@tnt.uni-hannover.de \\
      \addr Institute of Artificial Intelligence \\
      Leibniz University Hannover
      \UND      
      \name Bernd Bischl \email Bernd.Bischl@stat.uni-muenchen.de \\
      \addr Institute of Statistics, Munich Center for Machine Learning (MCML) \\
        Ludwig-Maximilians-Universität München}

\def\month{MM}  % Insert correct month for camera-ready version
\def\year{YYYY} % Insert correct year for camera-ready version
\def\openreview{\url{https://openreview.net/forum?id=XXXX}} % Insert correct link to OpenReview for camera-ready version

\begin{document}

\maketitle

\begin{abstract}
    Despite all the benefits of automated hyperparameter optimization (HPO), most modern HPO algorithms are black-boxes themselves. This makes it difficult to understand the decision process which leads to the selected configuration, reduces trust in HPO, and thus hinders its broad adoption. 
    Here, we study the combination of HPO with interpretable machine learning (IML) methods such as partial dependence plots.
    These techniques are more and more used to explain the marginal effect of hyperparameters on the black-box cost function or to quantify the importance of hyperparameters. 
    However, if such methods are naively applied to the experimental data of the HPO process in a post-hoc manner, the underlying sampling bias of the optimizer can distort interpretations. 
    We propose a modified HPO method which efficiently balances the search for the global optimum w.r.t. predictive performance \emph{and} the reliable estimation of IML explanations of an underlying black-box function by coupling Bayesian optimization and Bayesian Algorithm Execution. 
    On benchmark cases of both synthetic objectives and HPO of a neural network, we demonstrate that our method returns more reliable explanations of the underlying black-box without a loss of optimization performance. 
\end{abstract}

\section{Introduction}
\label{sec:introduction}

 % CONTEXT
The performance of machine learning (ML) models usually depends on many decisions, such as the choice of a learning algorithm and its hyperparameter configurations. Manually reaching these decisions is usually a tedious trial-and-error process. Automated machine learning (AutoML), e.g., hyperparameter optimization (HPO), can support developers and researchers in this regard.
By framing these decisions as an optimization problem and solving them using efficient black-box optimizers such as Bayesian Optimization (BO), HPO is demonstrably more efficient than manual tuning, and grid or random search \citep{bergstra2011algorithms, snoek2012practical,TurnerEMKLXG20,bischl2021hpo}. 
However, there is still a lack of confidence in AutoML systems and a reluctance to trust the returned best configuration~\citep{drozdal2020automl}. One reason for why some practitioners still today prefer manual tuning over automated HPO is that existing systems lack the ability to convey an understanding of hyperparameter importance and how certain hyperparameters affect model performance~\citep{hasebrook22whymanualtuning}, helping them to understand why a final configuration was chosen. 
% However, due to the lack of insight, there is a lack of confidence in AutoML systems and a reluctance to trust the returned best configuration~\citep{drozdal2020automl}.
% Some practitioners even prefer manual tuning because they feel they understand the process better \citep{hasebrook22whymanualtuning}.

% PROBLEM 
Desirable insights into hyperparameter effects or importance could in principle be generated by applying methods of interpretable machine learning (IML) to experimental data from the HPO process, 
specifically the final surrogate model generated by BO based on this HPO-data.
%Examples include marginal effects of individual hyperparameters or their importance. 
However, these methods -- even though possible from a technical perspective and used before~\citep{hutter14fanova, van2018hyperparameter, young18hyperspace, head2022scikit-optimize} -- should be used with caution in this context. The main reason is a sampling bias caused by the desire for efficient optimization during HPO~\citep{moosbauer2021xautoml}: Efficient optimizers typically sample more configurations in promising regions with potentially well-performing hyperparameter configurations, while other regions are underrepresented. This sampling bias introduces a surrogate-model bias in under-explored regions as the surrogate model is subject to high uncertainty in these regions. Consequently, explanations of HPO runs, such as partial dependence plots (PDPs) \citep{friedman2001greedy}, can be misleading as they also rely on artificially created evaluations in under-explored regions. \citet{moosbauer2021xautoml} had been the first to address this issue, and had proposed an approach to identify well-explored, rather small subregions in which PDPs can be estimated accurately. While this is  valuable, it still does not allow to accurately estimate hyperparameter effects globally.  

% SOLUTION
To anticipate these unintended effects of this sampling bias as effectively as possible already during the HPO process, we propose a modified BO algorithm that efficiently searches for the global optimum \emph{and} accurate IML estimates of the underlying black-box function at the same time. We build on the concept of Bayesian Algorithm Execution (BAX) \citep{neiswanger2021bax} to estimate the expected information gain (EIG) \citep{lindley56eig} of configurations w.r.t. the output of an interpretation method.
We ultimately couple BO with BAX and propose BOBAX as an efficient method that searches for accurate interpretations without a relevant loss of optimization performance. Our proposed method is generic as it is applicable to any BO variant (e.g., different acquisition functions or probabilistic surrogate models).
As IML technique we focus on PDPs \citep{friedman2001greedy}, which estimate the marginal effect(s) of features (in our case: hyperparameters) on the output by visualizing a marginal 1D or 2D function. 
PDPs constitute an established IML technique \citep{lemmens2006bagging, cutler2007random, wenger2012assessing, zhang2018opening}, have been in use for more than 20 years to analyze ML models, and have recently gained further interest in IML and XAI, and are also increasingly used to analyze hyperparameter effects in HPO and AutoML \citep{young18hyperspace, zela18autodl, head2022scikit-optimize}. 
We point out that our technique is in principle not limited to PDPs, but can be combined with any IML technique which can be quantitatively estimated from a surrogate model.

% EXPERIMENTS AND IMPACT 
In a benchmark study, we demonstrate how BOBAX consistently yields more reliable estimates for marginal effects estimated via the partial dependence method while maintaining the same level of optimization efficiency as commonly used methods. 
Finally, we demonstrate how BOBAX can give reliable insights into hyperparameter effects of a neural network during tuning yielding state-of-the-art performance. We believe that through our generic method, the potential of IML methods can be unlocked in the context of HPO, thus paving the way for more interpretability of and trust into human-centered HPO.
%\paragraph{Contributions}
Our contributions include:

\begin{enumerate}
    \item The direct optimization for an accurate estimation of IML statistics, e.g., marginal effects for single or multiple hyperparameters, as part of BO for HPO, making HPO interpretable and more trustworthy;
    \item The combination of BO and Bayesian Algorithm Execution (BAX), dubbed BOBAX, where BAX is used to guide the search towards more accurate estimation of IML statistics;
    \item Thorough study of different variants of BOBAX and baselines on synthetic functions; and
    % to show the interactions between both; and
    \item Empirical evidence that budget allocation regarding IML estimates does not come at the expense of significantly reduced optimization performance on a deep learning HPO benchmark.
\end{enumerate}

\section{Background}
\label{sec:background}

In this section, we formalize HPO and BO as the context of our work. We also give an overview of Bayesian Algorithm Execution (BAX) as it serves as basis for our work. 
\paragraph{Hyperparameter Optimization} The aim of HPO is to efficiently find a well-performing configuration of a learning algorithm. HPO is therefore commonly formalized as finding the minimizer $\lambdab^\ast \in \argmin\nolimits_{\lambdab \in \Lambda} c(\lambdab)$ of a \emph{black-box} cost function $c: \Lambda \to \R$ which maps a hyperparameter configuration $\mbox{$\lambdab = \left(\lambda_1, ..., \lambda_d\right)$} \in \Lambda$ to the validation error of the model trained by a learning algorithm run using $\lambdab$. The hyperparameter space $\Lambda = \Lambda_1 \times ... \times \Lambda_d$ can be mixed, containing categorical and continuous hyperparameters. Particularly in the context of AutoML, where whole machine learning pipeline configurations are optimized over, $\Lambda$ may even contain hierarchical dependencies between hyperparameters~\citep{ThorntonHHL13,OlsonM16}. 

% Introduce Bayesian Optimization
\paragraph{Bayesian Optimization} 
BO is a black-box optimization algorithm which has become increasingly popular in the context of HPO \citep{jones1998global, snoek2012practical}. 
BO sequentially chooses configurations $\lambdab^{(1)}, ..., \lambdab^{(T)}$ that are evaluated $c_{\lambdab^{(1)}}, ..., c_{\lambdab^{(T)}}$ to obtain an archive \mbox{$A_T = \left\{\left(\lambdab^{(i)}, c_{\lambdab^{(i)}}\right)\right\}_{i = 1, ..., T}$}. To choose the next configuration $\lambdab^{(T + 1)}$ as efficiently as possible, a surrogate model $\hat c$ is estimated on the archive $A_T$, and a new point is proposed based on an acquisition function that leverages information from the surrogate model $\hat c$. Typically, we chose a probabilistic model and estimate a distribution over $c$, denoted by $p(c~|~A_T)$. A common choice are Gaussian processes $c \sim \mathcal{GP}\left(\mu, k\right)$, characterized by a mean function $\mu: \Lambda \to \R$ and a covariance function $k: \Lambda \times \Lambda \to \R$. Acquisition functions usually trade off exploration (i.e., sampling in regions with few data points and high posterior uncertainty) and exploitation (i.e., sampling in regions with low mean). Common examples are the expected improvement (EI) \citep{jones1998global}, the lower confidence bound (LCB) \citep{jones2001taxonomy,SrinivasKKS10}, entropy search \citep{henning2012entropy, hernandez2014pes} and knowledge gradient~\citep{WuPWF17}. 

\paragraph{Marginal Effects of Hyperparameters} Practitioners of HPO are often interested in whether and how individual hyperparameters affect model performance. Not only is there a desire to gain model comprehension \cite{hasebrook22whymanualtuning}, also such insights can influence decisions, for example whether to tune a hyperparameter or not \citep{probst2018tunability}, or modify hyperparameter ranges. One interpretation measure that the community is looking at \citep{hutter14fanova,zela18autodl,young18hyperspace,van2018hyperparameter,abs-2202-11954} is the marginal effect of one or multiple hyperparameters $\lambdab_S$, $S \subset \{1, 2, ..., d\}$ on model performance, which is defined as\footnote{To keep notation simple, we denote $c(\lambdab)$ as a function of two arguments $(\lambdab_S, \lambdab_R)$ to differentiate components in the index set $S$ from those in the complement $R = \{1, 2, ..., d\} \setminus S$. The integral shall be understood as a multiple integral of $c$ where $\lambdab_j$, $j \in R$, are integrated out. }
\begin{eqnarray}
       c_{S}(\lambdab_S) := \E_{\lambdab_R}\left[c(\lambdab)\right]=\int_{\Lambda_R} c(\lambdab_S, \lambdab_R)~\textrm{d}\mathbb{P}(\lambdab_R).
       \label{eq:pdp}
\end{eqnarray}

In the context of HPO, $\P$ is typically assumed to be the uniform distribution over $\Lambda_R$ since we are interested in how hyperparameter values $\lambdab_S$ impact model performance uniformly across the hyperparameter space~\citep{hutter14fanova, moosbauer2021xautoml}. Since computing Eq. \eqref{eq:pdp} analytically is usually possible, the PDP method \citep{friedman2001greedy} approximates the integral, as in Eq. \eqref{eq:pdp}, by Monte Carlo approximation. 

\paragraph{Information-based Bayesian Algorithm Execution}

Information-based Bayesian Algorithm Execution (BAX) extends the idea of using entropy search for estimating global optima to estimating other properties of a function $f: \mathcal{X} \to \R$ \citep{neiswanger2021bax}. Similar to BO, BAX tries to sequentially choose points $\mathbf{x}^{(i)} \in \mathcal{X}$ in order to estimate the quantity of interest accurately with as few evaluations as possible. It is assumed that the quantity of interest can be computed as the output $\mathcal{O}_\mathcal{A}:= \mathcal{O}_{\mathcal{A}}(f)$ of running an algorithm $\mathcal{A}$ on $f$, e.g. top-k estimation on a finite set, computing level sets or finding shortest paths. 

Similarly to BO, BAX sequentially builds a probabilistic model $p(f~|~A_T)$, e.g., a GP, over an archive of evaluated points $A_T$. Based on $p(f~|~A_T)$, they derive the posterior distribution over the algorithm output $p(\mathcal{O}_\mathcal{A}~|~A_T)$. To build the archive $A_T$ as efficiently as possible, they choose to evaluate the point $\mathbf{x}^{(T+1)}$ which maximizes the expected information gain about the algorithm output $\mathcal{O}_\mathcal{A}$

\begin{eqnarray}
      \text{EIG}_T(\mathbf{x}) &:= \mathbb{H}\left[\mathcal{O}_\mathcal{A}|A_T\right] -\mathbb{E}_{f_{\mathbf{x}}|A_T}\left[\mathbb{H}\left[\mathcal{O}_\mathcal{A}|A_{T+1}\right]\right], 
    \label{eq:eig}
\end{eqnarray}

where $\mathbb{H}$ denotes the entropy, and $A_{T+1} := A_T \cup \left\{\left(\mathbf{x}, f_{\mathbf{x}}\right)\right\}$ with $f_{\mathbf{x}}$ the (unrevealed) value of $f$ at $\xv$. 
% Here, $f_{\mathbf{x}}$ denotes the (unrevealed) value of $f$ at ${\mathbf{x}}$, and $\mathbb{H}$ denotes the entropy. 

\citet{neiswanger2021bax} propose an acquisition function to approximate the \textrm{EIG} in Eq.~\eqref{eq:eig}. In its simplest form, the algorithm output $\mathcal{O}_\mathcal{A}$ in the \textrm{EIG} is replaced by the algorithm's execution path $e_\mathcal{A}$, i.e., the sequence of all evaluations the algorithm $\mathcal{A}$ traverses, which thus gives full information about the output. The expected information gain estimated based on the execution path $e_\mathcal{A}$ is given by

\begin{equation}
\begin{aligned}
      \text{EIG}_T^e(\mathbf{x}) &= 
      \mathbb{H}\left[e_\mathcal{A}|A_T\right] - \mathbb{E}_{f_{\mathbf{x}}|A_T}\left[\mathbb{H}\left[e_\mathcal{A}|A_{T+1}\right]\right] \\
      &= \mathbb{H}\left[f_{\mathbf{x}}|A_T\right] -\mathbb{E}_{e_{\mathcal{A}}|A_T}\left[\mathbb{H}\left[f_{\mathbf{x}}|A_T, e_\mathcal{A}\right]\right]. 
      \label{eq:eige}
\end{aligned}
\end{equation}

%% TODO: \mathbb{E}_{f_{\mathbf{x}} = \mathbb{E}_{p(f_{\mathbf{x}}|A_T)}
%% TODO: \mathbb{E}_{e_{\mathcal{A}}} = \mathbb{E}_{p(e_{\mathcal{A}}|A_T)}

where they used the symmetry of the mutual information to come up with the latter expression. The first term $\mathbb{H}\left[f_{\mathbf{x}}|A_T\right]$ is the entropy of the posterior predictive distribution at an input $\mathbf{x}$ and can be computed in closed form. The second term can be estimated as follows: A number of $n_\text{path}$ samples $\tilde f \sim p(f~|~A_T)$ is drawn from the posterior process. The algorithm $\mathcal{A}$ is run on each of the samples $\tilde f$ to produce sample execution paths $\tilde e_\mathcal{A}$, yielding samples $\tilde e_\mathcal{A} \sim p(e_\mathcal{A}~|~A_T)$, used to estimate the second term as described by \citet{neiswanger2021bax}.

\section{Related Work}
\label{sec:related}

Interpretability in AutoML refers either to (1) the interpretation of the resulting model returned by an AutoML system~\citep{xanthopoulos:2020, binder2020mosmafs, zachariah21interpretable, coors2021compboost}, or (2) the interpretation of hyperparameter effects and importance~\citep{moosbauer2021xautoml}. 
We focus on the latter, specifically the construction of accurate and unbiased estimators for, e.g., hyperparameter effects in HPO.
%contributing to more transparency, understanding and trust in AutoML~\citep{drozdal2020automl}.

%The desire for interpretable models (1) is driven by the need to explain causes for decisions in different application areas (e.g., in credit risk management). Therefore, it is of interest to develop HPO methods and AutoML systems that seek or prefer interpretable models \citep{binder2020mosmafs, pfisterer2019AutoML, xanthopoulos:2020, coors2021compboost}. This is undoubtedly a very important line of research, but it is outside the scope of this paper. 

%We focus on (2): gaining understanding by interpreting the inner workings of the HPO process itself. The drive to improve the explainability of HPO is driven by the fact that explainability and transparency promote trust in a system \citep{drozdal2020automl}. Furthermore, the insights gained into the process open up the possibility to work interactively with AutoML systems or to draw meaningful conclusions that contribute to the development of smarter AutoML systems in the longer term. 

There are HPO and AutoML frameworks that provide visualisations and interpretability statistics as additional outputs, e.g., \emph{Google Vizier} \citep{golovin2017googlevizier} and \emph{xAutoML}~\citep{abs-2202-11954} provide an interactive dashboard visualizing the progress of the optimization and insights via parallel coordinate plots and multi-dimensional scaling on the optimizer footprint. Similarly, the HPO frameworks \emph{optuna} \citep{akiba2019optuna} or \emph{scikit-optimize} \citep{head2022scikit-optimize} allow for quick and simple visualization of optimization progress and results. However, such relatively simple visualizations do not give a deeper understanding of which hyperparameter influence model performance in what way. 
% provide deeper insights into the HPO process. %, e.g., they do not help in understanding the effect or importance of hyperparameters. 

In the context of HPO, practitioners are commonly interested on the marginal effects of hyperparameters on model performance~\cite{hutter14fanova, young18hyperspace, zela18autodl} or the importance of hyperparameters on model performance \citep{hutter14fanova, biedenkapp2017ablation, van2018hyperparameter, probst2018tunability}. The latter is often directly derived from marginal effects of hyperparameters \citep{hutter14fanova}. Established HPO frameworks~\citep{head2022scikit-optimize, akiba2019optuna} as well as visualization toolboxes~\cite{abs-2202-11954} already make implementations of these methods accessible to users, however they neither discuss nor address a distortion of those arising due to a sampling bias. 
% In the context of HPO, hyperparameter importance and hyperparameter effects are particularly informative and suitable for drawing meaningful conclusions, e.g. hyperparameter importance via functional ANOVA \citep{hutter14fanova}, ablation studies \citep{biedenkapp2017ablation}, or tunability~\citep{probst2018tunability}. Hyperparameter effects are still less established as statistic with exceptions of \citep{hutter14fanova}, even though they contain more information than an importance score. 
While all of these approaches have their merits, none of them address the imprecision in the estimates of these interpretive measures caused by sample bias that is present in the archive sampled by BO, since BO tends to exploit promising regions while leaving other regions unexplored. 
So far, only \citet{moosbauer2021xautoml} explicitly proposed a post-hoc method that is able to identify subspaces of the configuration space in which accurate and unbiased PDPs can be computed. However, the method does not provide more accurate global IML estimates. To our knowledge, we are the first to propose a method that improves the sampling process of HPO to provide more accurate global estimates of such IML methods. 

%All of the above methods are post-hoc methods, i.e., they are applied to experimental data produced by an HPO run, but they are not designed to guide the optimization process in order to enhance interpretability. To our knowledge, we are the first to propose an ex-ante method, i.e., a method that adapts the sampling process of HPO to yield reliable explanations instead of correcting for a sampling bias in a post-hoc manner.

\section{BOBAX: Enhanced Estimation of Interpretability Measures for HPO}
\label{sec:methods}

We present our main contribution: BOBAX that efficiently searches for accurate marginal effect estimates of hyperparameters while maintaining competitive HPO performance. 
%
% First, we derive the expected information gain (EIG) with regards to the output of the partial dependence as IML method to estimate marginal effects. We then integrate the EIG into an efficient BO optimization procedure. We also elaborate on different theoretical and practical considerations of our method such as theoretic runtime complexity. We finally discuss how the partial dependence could be easily replaced or extended to other outputs of interest. 

\subsection{Expected Information Gain for Partial Dependence}

% Introduction -- WHY 
We first derive the information gained with regards to the estimate of a marginal effect of a hyperparameter $\lambdab_S$ \emph{if} we observe performance $c_{\lambdab^{(T+1)}}$ for a hyperparameter configuration $\lambdab^{(T+1)}$.  
To this end, we quantify and analyze how a marginal effect is estimated in the context of HPO. Two types of approximations are performed: First, instead of estimating the marginal effect with regards to the \textit{true}, but unknown and expensive objective $c$, we estimate the marginal effect of the surrogate model $\hat c$ \footnote{Constructed by BO, usually this will be the final surrogate model of the BO run, but this can also be applied interactively to intermediate models}, with $\hat c$ denoting the posterior mean of a probabilistic model $p(c~|~A_T)$. Secondly, we use the partial dependence method~\citep{friedman2001greedy} for efficient estimation of marginal effects of $\hat c: \Lambda \to \R$, which estimates Eq. \eqref{eq:pdp} by Monte-Carlo sampling: 

\begin{eqnarray}
    \varphi_{\lambdab_S} = \frac{1}{n} \sum_{i = 1}^n \hat c\left(\lambdab_S, \lambdab_R^{(i)}\right),
    \label{eq:pdpest}
\end{eqnarray}

with $\lambdab_S$ fixed and $\lambdab_R^{(i)} \overset{i.i.d.}{\sim} \P(\lambdab_R)$ a Monte-Carlo sample drawn from a uniform distribution $\P$. To bound the computational effort to compute the PDP, Eq. \eqref{eq:pdpest} is evaluated for a (typically equidistant) set of grid points $\{\lambdab_S^{(j)}\}_{j = 1, ..., G}$. The PDP visualizes $\varphi_{\lambdab_S}$ against $\lambdab_S$. 

To define the expected information gain for partial dependence $\textrm{EIG}_\textrm{PDP}$, we have the partial dependence method in terms of a formal execution path (see also Algorithm \ref{alg:pdp}): We iterate over all grid points, and compute the mean prediction $\hat c^{(g, i)}$. The execution path $e_\mathcal{A}$ thus corresponds to the Cartesian product $\left(\lambdab_S^{(g)}, \lambdab_R^{(i)}\right)$ for $g \in \{1, ..., G\}$ and $i \in \{1, ..., n\}$ of all grid points $\lambdab_S^{(g)}$ and the Monte-Carlo samples $\lambdab_R^{(i)}$. 

% Write down the Expected information gain
As proposed by \citet{neiswanger2021bax} as one variant, we estimate the information gained with regards to the execution path of $e_\mathcal{A}$ instead of estimating the execution path with regards to the algorithm output $O_\mathcal{A}$. Note that \citet{neiswanger2021bax} argued that the criterion in Eq. \eqref{eq:eige} is in general suboptimal, if for example large parts of the execution path $e_\mathcal{A}$ do not have an influence on the algorithm output. We argue, however, that it is not applicable to our use-case since every element in the execution path of the PD method contributes with equal weight to the computation of the partial dependence. Figure \ref{fig:branin_illustration} illustrates the computation of the PD based on the execution path, as well as the computation of the $\textrm{EIG}_\textrm{PDP}$.

\begin{minipage}{.4\textwidth}
\vspace*{-5cm}
\begin{small}
\begin{algorithm}[H]
    \begin{algorithmic}
        \STATE \textbf{Input } $G$, $\hat c$, $\left(\lambdab_R^{(i)}\right)\overset{i.i.d.}{\sim} \P(\lambdab_R)$\vspace{0.2cm}
        \STATE $\left(\lambdab_S^{(1)}, ..., \lambdab_S^{(G)}\right) \leftarrow$ equidist. grid on $\Lambda_S$
        \FOR{$g \in \{1,2,...,G\}$}
        \FOR{$i \in \{1,2,...,n\}$}
        \STATE $\lambdab^{(g, i)} \leftarrow \left(\lambdab_S^{(g)}, \lambdab_R^{(i)}\right)$
        \STATE $\hat c^{(g, i)} \leftarrow \hat c\left(\lambdab^{(g, i)}\right)$
        \STATE $e_\mathcal{A} \leftarrow e_\mathcal{A} \cup \left(\lambdab^{(g, i)}, \hat c^{(g, i)}\right)$
        \ENDFOR
        \STATE $ \varphi_{\lambdab_S^{(g)}} \leftarrow \frac{1}{n} \sum_{i = 1}^n \hat c^{(g, i)}$
        \ENDFOR \vspace{0.2cm}
        \STATE \textbf{Return } $\left(\lambdab_S^{(g)}, \varphi_{\lambdab_S^{(g)}}\right)$, $g = 1, ..., G$
    \end{algorithmic}
    \caption{PD algorithm with explicit execution path $e_A$}
    \label{alg:pdp}
\end{algorithm}
\end{small}
\end{minipage} \quad %
\begin{minipage}[t]{0.55\textwidth}
  \centering
  \includegraphics[width=0.7\textwidth]{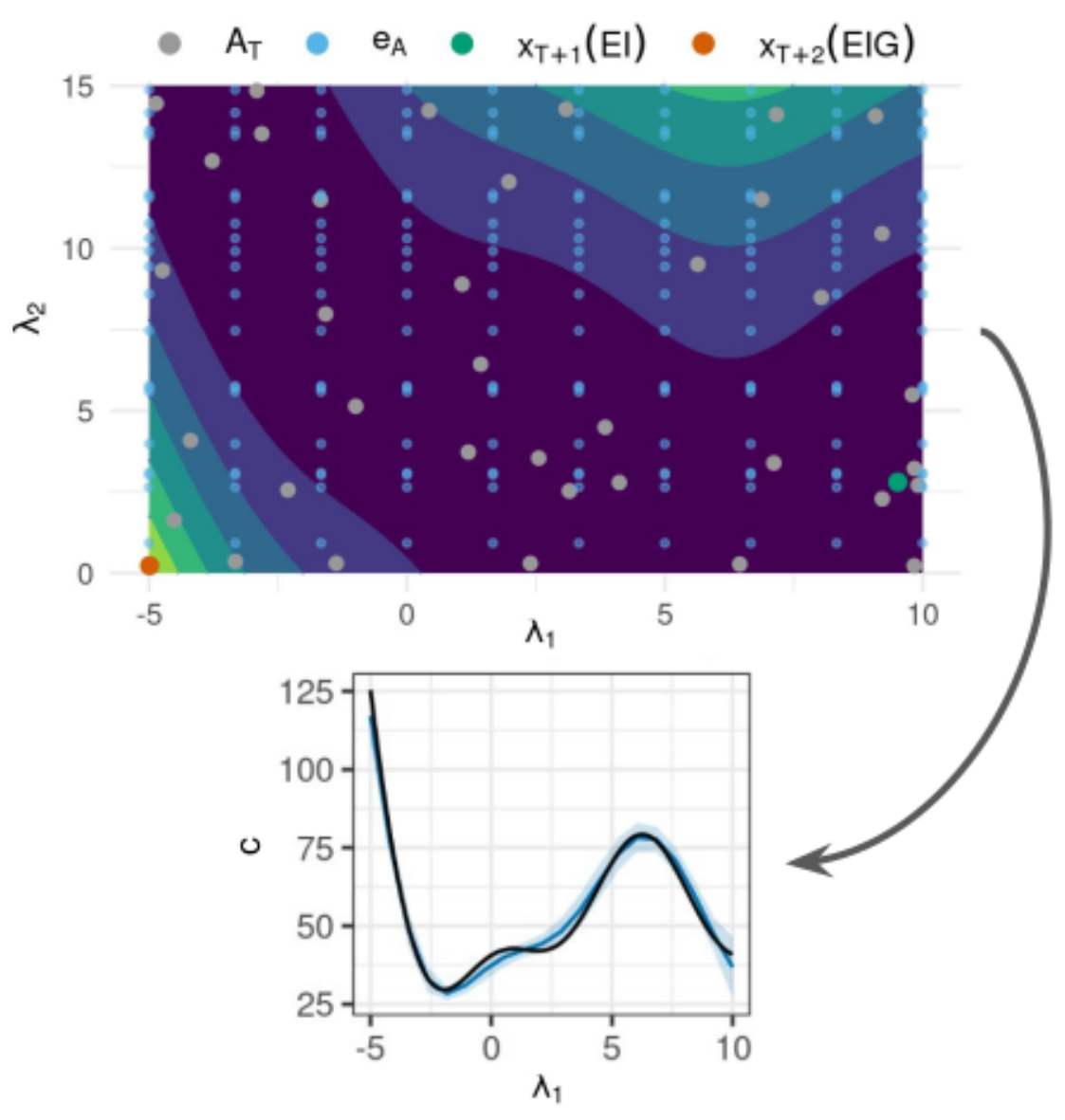}
  \captionof{figure}{Shown are the elements of $e_\mathcal{A}$ (blue)  PD method. The grey points show the configurations in the archive which are used by BO to construct the surrogate model. The green configuration is sampled by EI (showing more exploitation) while the orange point is the point maximizing the information gained about the PD estimate.} \label{fig:branin_illustration}
\end{minipage}

\subsection{BOBAX: Efficient Optimization and Search for More Accurate Interpretability Measures}
\label{ssec:bobax}
Given the $\textrm{EIG}_\textrm{PDP}$ for PD, the optimization for interpretability of hyperparameter effects as part of BO is possible by using the $\textrm{EIG}_\textrm{PDP}$ as acquisition function. However, interpretability alone is rarely of primary interest in practice; rather, the goal is to identify well-performing configurations and obtaining reasonable interpretations at the same time. We propose a method, dubbed BOBAX, that allows to efficiently search for explanations without a relevant loss of optimization efficiency.  

BOBAX is an interleaving strategy which performs BO, and iterates between using the EI (or any other suited acquisition function) and the $\textrm{EIG}_\textrm{PDP}$ as acquisition function. Although we have investigated also more complex variants (see Appendix \ref{app:bobax_variants}), interleaving $\textrm{EIG}_\textrm{PDP}$ in every $k$-th iteration is simple yet efficient. The smaller $k$ is, the higher is the weight of optimizing for accurate interpretations in a BO run. We note that this strategy can replace other interleaving exploration strategies, such as random samples~\citep{hutter2011smac}, since optimizing for interpretability can be seen as another strategy to cover the entire space in an efficient manner.\footnote{One might have also considered addressing this as a multi-objective problem since we have two objectives: (i) finding the optimum and (ii) obtaining good PDPs. However, usually post-hoc multi-objective optimizers construct a Pareto front of a set of multiple candidate solution, which we are not interested in here. Instead, in each iteration of BO, the optimizer has to choose a concrete trade-off between both objectives. For dynamically balancing out this trade-off, please also refer to the next section.}
%without caring about the approximation quality of the Pareto front. 
%Although there might be other trade-offs than we chose (i.e., the extrema of the trade-offs), our results indicate that our goal was achieved by providing better PDPs without losing optimization quality.}

%Since EIG is some type of exploration, which might be more favorable in the beginning of a BO run, annealing strategies to reduce the frequency of EIG use can be easily implemented by using state variables for $k$.  

From a practitioner's point of view, it may be reasonable to consider accuracy of interpretations rather as a constraint than an objective function to optimize for. As soon as this constraint is fulfilled, a user may want to invest all remaining budget into optimization only. Therefore, we also propose an adaptive variant of BOBAX, dubbed a-BOBAX, which performs the interleaving strategy in BOBAX as described above in a first phase, and transitions into optimization only in a second phase as soon as the constraint is fulfilled. To allow a user to input a meaningful constraint, the constraint must itself be interpretable by a user. Therefore, we define this constraint by a desired average width of confidence intervals around PD estimates, using the definition\footnote{Confidence intervals are defined as $\varphi_{\lambdab_S^{(g)}} \pm q_{1-\alpha/2} \cdot \hat s_{\lambdab_S^{(g)}}$ around the PD estimate. $\hat s_{\lambdab_S^{(g)}}$ denotes the uncertainty of a PD estimate for a grid point $g$. As default, we look at $\alpha=0.05$. } of \citet{moosbauer2021xautoml}. As an example, a user may want to specify a tolerance $\pm 1\%$ in validation accuracy in estimation of PDs (see green tolerance bands in Figure \ref{fig:example_learning_rate_lcbench} for illustration). 

\subsection{Theoretical and Practical Considerations}

\paragraph{Runtime Complexity}
\label{par:complexity}

Since BOBAX comes with additional overhead, we discuss this here in more detail. The computation of the expectation requires posterior samples of the execution path $e_\mathcal{A} ~ \sim p(e_\mathcal{A}~|~A_T)$. This is achieved by sampling from the posterior GP $\tilde c ~ \sim p(c~|~A_T)$ and execution of $O_\mathcal{A}$ on those samples, which may produce a computational overhead depending on the costs of running $O_\mathcal{A}$. We assume that executing $O_\mathcal{A}$ is neglectable in terms of runtime. However, to compute the entropy $\mathbb{H}\left[c_{\lambdab}|A_T, e_\mathcal{A}\right]$, the posterior process needs to be trained based on $A_T \cup e_\mathcal{A}$ (which has size $T + n \cdot G$). Thus, the overall runtime complexity is dominated by $\order\left(n_\text{path} \cdot (T + n)^3\right)$, as we compute the entropy $n_\text{path}$ times to approximate the expectation and since training a GP is cubic in the number of data points. 
Therefore, we recommend to keep an eye on the runtime overhead of the calculation of $\textrm{EIG}_\textrm{PDP}$ in relation to evaluating $c$ (e.g., training and evaluating an ML algorithm). Especially in the context of deep learning, the evaluation of a single configuration is usually by orders of magnitude higher than that of computing the $\textrm{EIG}_\textrm{PDP}$\footnote{In our case, the computation of the $\textrm{EIG}_\textrm{PDP}$ was ranging from the order of a few seconds to a few minutes.}. Also, we would like to emphasize that the implementation of our method is based on GPflow \citep{matthews17gpflow}, which allows fast execution of GPs on GPUs. Since GPUs are typically in use for training in the context of DL anyway, they can easily be leveraged in between iterations to speed up the computation of the $\textrm{EIG}_\textrm{PDP}$.  

\paragraph{Marginal Effects for Multiple Hyperparameters}
\label{par:multiple_variables}

Until now we have assumed that a user specifies a single hyperparameter of interest $\lambdab_S$ for which we will compute the PD. However, it is difficult to prioritize the hyperparameter of interest a-priori. Fortunately, it is possible to extend the execution path to compute $\textrm{EIG}_\textrm{PDP}$ by the respective execution paths of the PDs with regards to all variables $e_\mathcal{A} = e_{\mathcal{A}, \lambdab_1} \cup e_{\mathcal{A}, \lambdab_2} \cup ... \cup e_{\mathcal{A}, \lambdab_d}$. We investigate the differences between $\textrm{EIG}_\textrm{PDP}$ for a single hyperparameter vs. for multiple hyperparameters in more detail in Appendix \ref{app:benchmark}; in the practical use-case (see Section \ref{sec:use_case}), we compute the $\textrm{EIG}_\textrm{PDP}$ for multiple hyperparameters.

\section{Benchmark}
\label{sec:benchmark}

In this section, we present experiments to demonstrate the validity of our method. In particular, we look at: 

\paragraph{Hypothesis H1} % (Optimization for More Accurate Interpretability Measures):] 
Performing BO with $\text{EIG}_\textrm{PDP}$ as acquisition function is more efficient than random search in optimizing for accurate interpretations
\paragraph{Hypothesis H2}
Through BOBAX the accuracy of marginal effect estimates is clearly improved without a significant loss of optimization performance.  

% Further hypotheses are investigated in Appendix \ref{app:benchmark}. 

\subsection{Experimental Setup}

\begin{figure*}[tbh]
    \centering
    \includegraphics[width=1\textwidth]{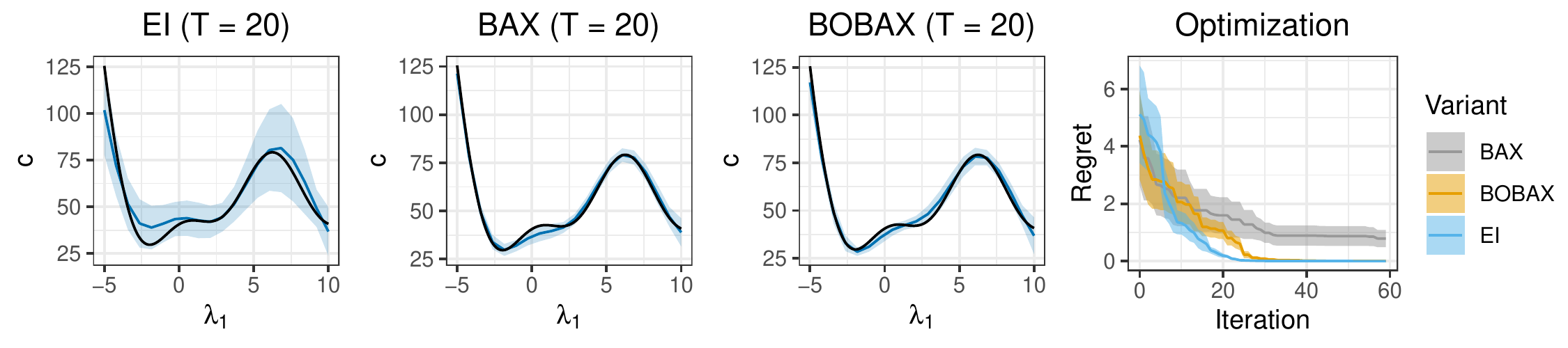}
    \caption{The first three plots show the estimated PD with $95\%$ confidence interval (blue) based on the surrogate model $\hat c$ after $T = 30$ iterations vs. the true marginal effect (black). BAX and BOBAX yield more accurate estimates for the PD as compared the BO with EI. The right plot shows the cumulative regret for the three methods. BAX, which is not performing optimization at all, is also clearly outperformed in optimization performance. BOBAX reaches the optimization result of BO with EI only after a few more iterations. }
    \label{fig:branin_example}
\end{figure*}

\paragraph{Objective Functions} We apply our method to synthetic functions which are treated as black-box function during optimization: Branin ($d = 2$), Camelback ($d = 2$), Stylinski-Tang ($d = 3$), Hartmann3 ($d = 3$) and Hartmann6 ($d = 6$).

\paragraph{Algorithms} To investigate H1, we consider BO with $\text{EIG}_\text{PDP}$ as acquisition function (\textbf{BAX}). For H2, we consider BOBAX as described in Algorithm \ref{alg:bobax}, where we iterate evenly ($k = 2$) between EI and $\text{EIG}_\text{PDP}$ as acquisition function. Following \citet{neiswanger2021bax} we set the number of execution path samples to $20$ to approximate the expectation in Eq. \eqref{eq:eige} in both variants. As strong baseline for accurate PDs we consider random search (\textbf{RS}) and BO with posterior variance as acquistion function (\textbf{PVAR}) as a pure exploration case of LCB. As strong baseline for optimization we consider BO with EI (\textbf{BO-EI}). Further variants of our methods (e.g., different frequencies of interleaving) and additional baselines (such as BO with LCB with different exploration factors, or BO with EI and random interleaving) are described in Appendix \ref{app:benchmark}.

\paragraph{Evaluation} We evaluate the accuracy of PD estimates by comparing the PD $\varphi_S^{(g)}$ (estimated based on $\hat c$) against the PD $\tilde \varphi_S^{(g)}$ computed on the ground-truth objective function $c$, approximated with the same sample $\lambdab_R^{(i)}$ and the same grid size $G$. As measure we use the $L_1$ distance $\textrm{d}_{\text{L}_1} := \frac{1}{G}\sum_{g = 1}^G \left|\varphi_S^{(g)} - \tilde \varphi_S^{(g)}\right|$ averaged over all grid points. 
%Additionally, we report the mean confidence (MC) of a PD estimate as $\textrm{d}_{\text{L}_1} := \frac{1}{G}\sum_{g = 1}^G \hat s(\lambdab_S^{(g)})$ to evaluate the confidence in the estimate. 
To assess optimization performance, we report the simple regret $c(\hat \lambdab) - c(\lambdab^\ast)$, where $\lambdab^\ast$ denotes the theoretical optimum of a function, and $\hat \lambdab \in \textrm{argmin} \left\{c_{\lambdab} ~|~ (\lambdab, c_{\lambdab}) \in A_T\right\}$ is the best found configuration during optimization.

\paragraph{Further Configurations} A Gaussian process with a squared exponential kernel is used as surrogate model for all BO variants, and PDs are estimated on the respective surrogate models. 
For RS, a GP (with same configuration) is fitted on $A_T$ and the PD is computed thereon. Acquisition function optimization is performed by randomly sampling $1500$ configurations, evaluating the respective acquisition function and returning the best. Each (BO) run is given a maximum number of $30 \cdot d$ function evaluations.

\paragraph{Reproducibility and Open Science} The implementation of methods as well as reproducible scripts for all experiments are publicly made available. Each experiment is replicated $20$ times based $20$ different seeds fixed across all variants. More details on the code and on computational can be found in Appendix \ref{app:code}.

\subsection{H1: More accurate interpretations}
\label{ssec:exp_reliable_pdp}

% First, we observe that BAX yields better explanations
Our experiments support hypothesis H1, i.e., we can achieve more accurate PD estimates more efficiently through targeted sampling via the $\textrm{EIG}_\textrm{PDP}$. An example run on the Branin function shown in Figure \ref{fig:branin_example} illustrates the behavior of the methods that is observable across all experiments: BAX is yielding clearly more accurate PDPs than BO with EI already after few iterations.  Figure \ref{fig:ranks} in Appendix \ref{app:benchmark_results} supports that PDs estimated on data produced by BO with EI might provide not only quantitatively, but also qualitatively wrong information in terms of ranking the values $\varphi_{\lambdab_S^{(g)}}$ differently than the ground-truth. As expected, increased accuracy of interpretations through BAX comes to the cost of optimization efficiency.
Results aggregated across all problems and replications confirm this behavior on a broader scale, see Table \ref{tab:part1_rank_table}\footnote{We note that the different functions live on different scales s.t. we normalized it by showing relative metrics wrt baselines, such RS for PDP estimates and EI for optimization regret.}. BAX is producing more accurate PDPs than RS (which can be assumed to converge against the true marginal effect) already at early stages, and is strongly significantly ($\alpha = 1 \%$) outperforming RS with less iterations. We conclude that both BAX and PVAR can contribute to approximating the true marginal effect well, but BAX is converging faster. In addition, BO with EI is significantly outperformed in terms of accuracy of PDPs, which supports our assumption of lowered quality caused through a heavy sampling bias.

\begin{footnotesize}
\begin{table*}[tbph]
    \centering
    \caption{\textbf{Left}: $L1$ error of the estimated PDP w.r.t. the ground truth PDP, relative to RS as baseline. Negative values mean a relative reduction of the $L1$ error compared to random search. \textbf{Right:} Optimization relative to BO-EI as baseline. Results are averaged across all $20$ replications. Best values are bold, and  values are underlined if not significantly worse than the best based on a Post-Hoc Friedman test ($\alpha = 1 \%$), see also \cite{demsar06posthoc, garcia10friedmanaligned} and Appendix \ref{app:benchmark_details} for more details. }
    \vspace*{0.1cm} 
    \label{tab:part1_rank_table}
    \begin{tabular}{rrrrr}
        \toprule
         & \multicolumn{4}{c}{Relative $\textrm{d}_{L_1}(\textrm{PDP})$ after} \\ \cmidrule(l){2-5} 
         & 25\% & 50\% & 75\% & 100\% \\ 
         & \multicolumn{4}{c}{Max. iterations spent} \\ \midrule
        RS & 0.00 & 0.00 & 0.00 & \underline{0.00} \\
        BO-EI & 0.18 & 0.39 & 0.47 & 0.67 \\
        PVAR & 0.13 & \underline{-0.08} & \underline{0.08} & \underline{0.14} \\
        BAX & \underline{\textbf{-0.17}} & \underline{\textbf{-0.20}} & \underline{\textbf{-0.07}} & \textbf{\underline{0.00}} \\
        BOBAX & \underline{-0.14} & \underline{-0.16} & \underline{-0.04} & \underline{0.03} \\
        \bottomrule
    \end{tabular} \hspace{0.1cm}
    \begin{tabular}{rrrrr}
        \toprule
         & \multicolumn{4}{c}{Relative optimization regret after} \\ \cmidrule(l){2-5} 
         & 25\% & 50\% & 75\% & 100\% \\ 
         & \multicolumn{4}{c}{Max. iterations spent} \\ \midrule
        RS & 2.42 & 160.99 & 530.70 & 951.47 \\
        BO-EI & \textbf{\underline{0.00}} & \textbf{\underline{0.00}} & \textbf{\underline{0.00}} & \textbf{\underline{0.00}} \\
        PVAR & 3.38	& 232.14 & 741.69 & 1887.22 \\
        BAX & 2.27 & 242.062 & 602.15 & 1408.62 \\
        BOBAX & 1.68 & \underline{5.04} & \underline{4.73} & \underline{3.26} \\
        \bottomrule
    \end{tabular}     
\end{table*}
\end{footnotesize}

\subsection{H2: More accurate interpretations at no relevant loss of optimization efficiency}
\label{ssec:exp_bobax}

% First, we observe that BAX yields better explanations
Our experiments also support hypothesis H2, i.e., with BOBAX we can achieve clearly more accurate PD estimates while maintaining a competitive level of optimization efficiency. Table \ref{tab:part1_rank_table} compares the accuracy of PD estimates (measured via $\textrm{d}_{L_1}$) and optimization regret as compared to baselines RS and BO-EI, respectively, aggregated over all five objective functions. (BO)BAX allows for more accurate PDPs than the other methods, with diminishing relative distance to RS, while BO with EI is clearly outperformed. On the other hand, it can be observed that BOBAX is giving optimization performance comparable to BO with EI throughout the course of optimization, whereas RS is clearly outperformed. So, BOBAX combines the best of both worlds: good interpretability (even better than RS) and efficient optimization (on par with BO-EI). Figure \ref{fig:part2_boxplot} in Appendix \ref{app:benchmark_results} shows that this effect is visible for all objective function, but the strength of the effect depends on the objective functions. 

We conclude that our experiments support that BOBAX makes no (or only little) compromises in optimization performance, but yields clearly better estimates of marginal effects at the same time.

\section{Practical HPO Application}

\begin{figure}[thb]
    \centering
    \includegraphics[width=0.8\textwidth]{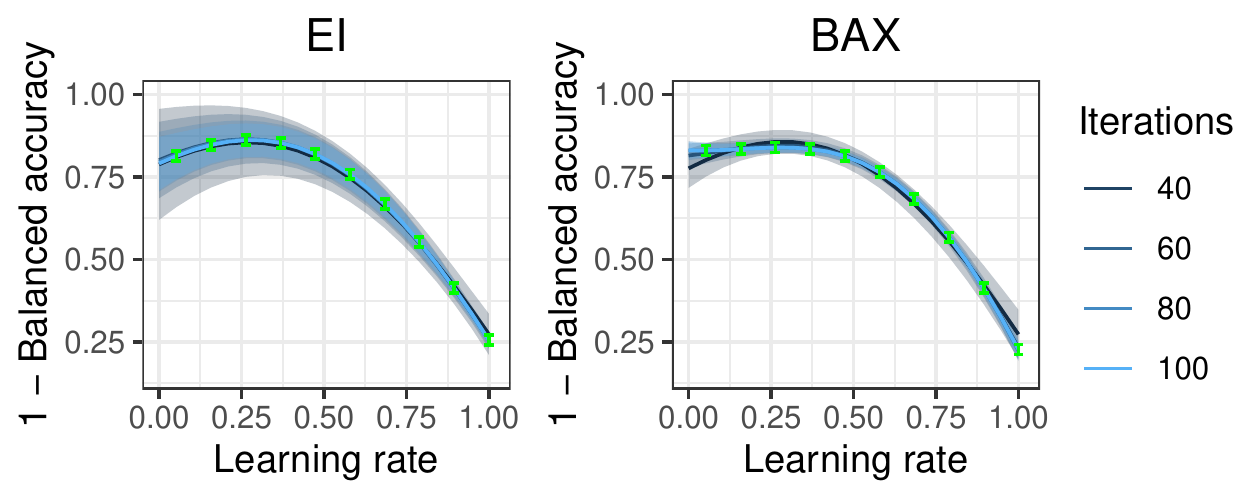}
    \caption{Comparing PDP evolution for number of iterations for EI and BAX. BAX returns fairly certain PDPs early on, whereas BO with EI requires much more time.}
    \label{fig:example_learning_rate_lcbench}
\end{figure}

\begin{table*}[]
    \centering
    \caption{Iterations needed to reach the desired precision of PD estimate of $\pm 1.5$ balanced accuracy points, accuracy of the final PD based on the L1 error to the ground truth, as well as the final model performance reached. Results are averaged across all 30 replications and all 15 datasets. Best values are bold, and  values are underlined if not significantly worse than the best based on a Post-Hoc Friedman test ($\alpha = 1 \%$), see also \cite{demsar06posthoc, garcia10friedmanaligned} and Appendix \ref{app:benchmark_details} for more details. }
    \begin{tabular}{rccc}
    \toprule
         & Iterations to desired precision & Rel. $\textrm{d}_{L_1}$ (PDP) & 1 - Balanced Accuracy \\ \midrule
         RS & 14.91 & \textbf{\underline{0.49}} & 22.56 \\
         BO-EI & 22.59 & 0.57 & \textbf{\underline{19.38}} \\
         BAX & 9.85 & 0.51 & 23.97 \\
        a-BOBAX & \textbf{11.56} & \underline{0.52} & \underline{19.94} \\ \bottomrule
    \end{tabular}
    \label{tab:lcbench}
\end{table*}

We demonstrate a-BOBAX on a concrete HPO scenario, following the setup of \citet{moosbauer2021xautoml}. We tune common hyperparameters of a neural network with regards to balanced validation accuracy on 15 different datasets respresenting different tasks from different domains (see Tables \ref{tab:searchspace}, \ref{tab:datasets} in Appendix \ref{app:practical_HPO}) using the interface provided by YAHPO gym \citep{pfisterer21yahpo}. We compare RS, EI, BAX, and adaptive BOBAX (a-BOBAX). In a-BOBAX, we set the desired width of confidence intervals to $\pm 1.5\%$ balanced accuracy points; we emphasize thought, that this value can be set by the user. % ; different values for this user-set threshold are studied in Appendix \ref{app:practical_HPO}. 
For a-BOBAX, we compute the $\textrm{EIG}_\textrm{PDP}$ jointly for the PDPs of \emph{learning rate}, \emph{dropout}, \emph{max. number of units}, \emph{weight decay}, and \emph{momentum}. The respective methods ran under the same conditions as in Section \ref{sec:benchmark}, but were replicated $30$ times. 

Figure \ref{fig:example_learning_rate_lcbench} shows how accuracy of the PD estimate increases over time for BO with EI vs. BAX. We observe that BAX is clearly more efficient in returning an accurate estimate, which is in line with the results we observed in Section \ref{sec:benchmark}. As motivated in Section \ref{ssec:bobax}, a practitioner might prefer to rather ensure a minimum accuracy of IML measures, and therefore, handle this rather as a constraint than as an objective. Table \ref{tab:lcbench} is showing the time to reach the desired precision of $\pm 1.5\%$ for the PDP, as well as final accuracy of PDs and final optimization performance, aggregated over all experiments an replications. We observe that a-BOBAX is (i) significantly faster in reaching the desired precision threshold, allowing a user to interact earlier with confidence, (ii) is comparable to RS in terms of final accurate representation of PDs, and (iii) comparable to BO-EI in terms of optimization performance. Note that the effect again depends strongly on the respective dataset (see Figures \ref{fig:detailed_results_lcbench1}, \ref{fig:detailed_results_lcbench2}, \ref{fig:detailed_results_lcbench3} in Appendix \ref{app:practical_HPO_additional_results}). 
%If even considering that with the help of a-BOBAX one could now intervene in the optimization process in an interactive scenario (e.g., changing tuning ranges of hyperparameters), one could by all means accept these (extremely small) losses. %In the example shown in Figure \ref{fig:example_learning_rate_lcbench}, a practitioner try to extend the range for the learning rate to potentially reduce validation accuracy further. 

\label{sec:use_case}

\section{Discussion and Conclusion}
\label{sec:conclusion}

% xxx

% xxx

\paragraph{Findings}
We proposed (adaptive) BOBAX, modifying Bayesian Optimization (BO) for black-box optimization and HPO to enhance interpretability of the optimization problem at hand. We achieved this by adapting BAX to optimize for accurate marginal effects and then interleaved BO and BAX. We further showed that BOBAX can significantly enhance the accuracy of PD estimates during an optimization procedure, while not losing optimization performance. 

\paragraph{Usage}
If a user has some desired precision of the IML estimates in mind, a-BOBAX allows them to make use of BAX only until this level is not reached yet and will focus on the optimization quality afterwards. This simple, yet efficient strategy allows to get the most out of the overall budget. 
%We believe that in particular an adaptive method, where a user can specify a certain precision in terms of desired width of a confidence interval, is fulfilled, until it transitions into optimization mode. 

\paragraph{Critical View and Limitations}
Even though the usage of EIG is beneficial to the quality of a PD estimate, there are also examples where no significant improvement is observed. We assume that this particularly holds for hyperparameters that have a simple (and therefore easy-to-learn) effect on performance. Consequently, the marginal effect is easily learned for any of the methods. In addition to using the adaptive version of BOBAX, we recommend dropping these simple-to-learn hyperparameters from the joint computation of the EIG (\ref{par:multiple_variables}) as soon as the PDPs are sufficiently certain.
% For the practical use of our method, we therefore recommend the following: (1) Use adaptive BOBAX to focus all budget on optimization once a desired precision is achieved, and (2) if a single hyperparameter effect has been learned up to a certain precision, exclude it from the joint computation of the EIG (Section \ref{ssec:multiple_variables}) to save computational overhead. 
Furthermore, our method comes at a computational overhead, being slightly larger than traditional BO since computing EIG with BAX costs a bit more compute time. In terms of application to HPO, we expect that the cost for training and validating hyperparameter configurations or architectures of neural networks will be much larger than BOBAX's overhead in most relevant cases.

%. We do not recommend to use our method in situations where Gaussian process prediction is more expensive that actually evaluating the objective $c$. In particular, we see our method in the context where model fitting is expensive. Particularly for deep learning, it can be emphasized that -- if we anyways have access to GPUs -- these can be leveraged in between iterations for speeding up the estimation of marginal effects. 

\paragraph{Outlook}
We believe that BOBAX will contribute in particular towards more human-centered HPO, where developers can start inspecting intermediate results as soon as desired confidence was reached and then adapt the configuration space if necessary. Although we focused on PDPs as an interpretability method, extending our BOBAX idea to other IML approaches would be straightforward and opens up new follow up directions. As one next step, we envision extending BOBAX to the multi-fidelity setting~\citep{LiJDRT17,FalknerKH18} which is required for more expensive HPO and AutoML problems. Last but not least, we emphasize that we developed BOBAX primarily for HPO problems, but it can also be applied to any black-box optimization problem, e.g., in engineering or chemistry.

% \section{Acknowledgments}
% This work has been partially supported by the German Federal Ministry of Education and Research (BMBF) under Grant No. 01IS18036A. The authors of this work take full responsibilities for its content.

%% If their work carries a significant risk of harm, authors are required to include a Statement of Broader Impact: 
% \subsubsection*{Broader Impact Statement}
% In this optional section, TMLR encourages authors to discuss possible repercussions of their work, notably any potential negative impact that a user of this research should be aware of. 
% Authors should consult the TMLR Ethics Guidelines available on the TMLR website
% for guidance on how to approach this subject.

%% Optional: 
% \subsubsection*{Author Contributions}
% If you'd like to, you may include a section for author contributions as is done
% in many journals. This is optional and at the discretion of the authors. Only add
% this information once your submission is accepted and deanonymized. 

\subsubsection*{Acknowledgments}
Use unnumbered third level headings for the acknowledgments. All
acknowledgments, including those to funding agencies, go at the end of the paper.
Only add this information once your submission is accepted and deanonymized. 

\newpage 

\bibliography{references}
\bibliographystyle{tmlr}

\newpage

\appendix
\section{Appendix}

\section{Additional methodological aspects}

\subsection{Interpretability methods beyond the PDP}
\label{app:methods_of_interpretability}

BOBAX is generic in the sense that it can be applied to other IML methods than the PDP that are of interest to the user, as long as a the execution path of the respective method is accessible to BOBAX.

While we considered the partial dependence method to estimate main effects (i.e., the marginal effect of a single hyperparameter $\lambdab_s$ on estimated performance) in our experiments, Algorithm \ref{alg:pdp} can be extended to estimate interaction effects of two hyperparameters $S = \{s, s^\prime\}$. This is done by simply replacing the grid points in Algorithm \ref{alg:pdp} by a two-dimensional grid $\left(\lambdab_s^{(g)}, \lambdab_{s^\prime}^{(g^\prime)}\right)$ for all pairs $g, g^\prime \in \{1, 2, ..., G\}$ with $\left(\lambdab_s^{(1)}, ..., \lambdab_s^{(G)}\right)$ and $\left(\lambdab_{s^\prime}^{(1)}, ..., \lambdab_{s^\prime}^{(G)}\right)$ representing equidistant grids. With this modified execution path our method is be straightforwardly applied to estimate interaction effects. 

Also, other methods within IML can be optimized for with BOBAX; for example, the hyperparameter importance via permutation feature importance (PFI) \citep{fisher2019pfi}. Importance of a single hyperparameter $\lambdab_S$ is computed by shuffling the the values of this hyperparameter in the $A_T$, resulting in a modified archive $\tilde A_{T, \lambdab_S}$ and the difference in errors of the model $\hat c$ on $A_T$ and on $\tilde A_{T, \lambdab_S}$ is compared. The respective execution path $e_\mathcal{A}$ is the joint set of all shuffled versions of the archive $\tilde A_{T, \lambdab_1} \cup \tilde A_{T, \lambdab_2} \cup ... \cup \tilde A_{T, \lambdab_d}$.  

\subsection{BOBAX and Variants}
\label{app:bobax_variants}

\begin{algorithm}[H]
    \caption{BOBAX} \label{alg:bobax}
    \begin{algorithmic}
        \STATE \textbf{Input } $k$, $n_\text{init}$, $O_\mathcal{A}$\vspace{0.2cm}
        \STATE $A_T \leftarrow$ Sample initial design of size $n_\text{init}$ over $\Lambda$
        \WHILE{stopping criterion not met}
            \IF{$T \mod k = 0$}
                \STATE $\lambdab^{(T+1)} \leftarrow \arg \max_{\lambdab \in \Lambda} \textrm{EIG}_\textrm{PDP}(\lambdab)$
            \ELSE
                \STATE $\lambdab^{(T+1)} \leftarrow \arg \max_{\lambdab \in \Lambda} \text{EI}(\lambdab)$
        \ENDIF
        \STATE $ c_{\lambdab^{(T+1)}} \leftarrow c(\lambdab^{(T+1)})$
        \STATE $A_{T+1} \leftarrow A_T \cup \{\left(\lambdab^{(T+1)}, c_{\lambdab^{(T+1)}}\right)\}$
        \STATE $T \leftarrow T + 1$
        \ENDWHILE \vspace{0.2cm}
        \STATE \textbf{Return } $A_T, O_\mathcal{A}(\hat c)$
    \end{algorithmic}
\end{algorithm}\vspace{0.2cm}

Algorithm \ref{alg:bobax} shows the BOBAX algorithm as introduced and discussed in the main paper. We have investigated two more alternative acquisition functions to trade-off interpretability and optimization efficiency. One is a probabilistic variant of interleaving $\textrm{EIG}_\textrm{PDP}$, where in every iteration

\begin{eqnarray*}
    \lambdab^{(T+1)} = \textrm{arg max}_{\lambdab \in \Lambda} \begin{cases}  \textrm{EIG}_\textrm{PDP}(\lambdab) & \textrm{if } p \le \pi \\
    \textrm{EI}(\lambdab) & \textrm{if } p > \pi
    \end{cases}
\end{eqnarray*}

where $p \sim \textrm{Unif}(0, 1)$ and $\pi$ is a threshold set by a user. If $\pi$ is set to $0.5$ this corresponds to the probabilistic variant of Algorithm \ref{alg:bobax} with $k = 2$. We call this variant $\textrm{BOBAX}_\textrm{prob}^\pi$. This method also opens up the possibility to reduce 
the relative amount search for interpretability (as a kind of exploration) over time by an annealing strategy where the probability $\pi$ is lowered over time. 

As a second variant, we investigated a multiplicative variant of $\textrm{EIG}_\textrm{PDP}$ and $\textrm{EI}$ inspired by \cite{hvarfner2022pibo}: 

\begin{eqnarray*}
    \textrm{EIBAX}^{\beta}(\lambdab) = \textrm{EI}(\lambdab) \cdot \textrm{EIG}_\textrm{PDP}(\lambdab)^{\beta / T}, 
\end{eqnarray*}

where the values of a sampled batch of $\textrm{EIG}_\textrm{PDP}(\lambdab)$ are min-max-scaled to $[0,1]$. Note that in comparison to the interleaving strategy, this method has a computational disadvantage since it requires to compute the $\textrm{EIG}_\textrm{PDP}$ in \emph{every} iteration. 

Note that in any of the variants above, the EI can be replaced by any other acquisition function. 

\section{Benchmark}
\label{app:benchmark}

\subsection{Additional Details}
\label{app:benchmark_details}

\paragraph{Details on evaluation}

We performed a statistical test to allow for conclusions as to whether the methods compared ($\textbf{RS}, \textbf{EI}, \textbf{BAX}, \textbf{BOBAX}$) are performing \textit{significantly} differently in terms of (1) quality of the PD estimate measured by $\textrm{d}_{\textrm{L}_1}$, (2) optimization performance as measured by regret in Table \ref{tab:part1_rank_table}. We applied a \emph{Friedman aligned ranks test} as described in \citep{garcia10friedmanaligned} on the respective performance values on different objective functions and replications to conclude whether there is a difference between methods. Note that the chosen test is recommended over the Friedman test by \cite{garcia10friedmanaligned} in particular if the number of algorithms is low (four to five) because of an increased power. We applied a post hoc test with \textit{Hommel} correction for multiple testing, and report statistical significance based on corrected p-values. We rely on the implementation \texttt{scmamp}\footnote{\url{https://github.com/b0rxa/scmamp}}.

\paragraph{Comparison with additional baselines}
\label{par:additional_baselines}

As additional baselines, we are running BO with LCB $\hat c(\lambdab) + \tau \cdot \hat s^2(\lambdab)$ acquisition function with different values of $\tau \in \{1, 2, 5\}$, denoted by $\textbf{LCB}^1$, $\textbf{LCB}^2$, $\textbf{LCB}^5$. Also, we are running BO with interleaved random configurations every $k \in \{2, 5, 10\}$ iterations, denoted by $\textbf{BO-RS}^2, \textbf{BO-RS}^5, \textbf{BO-RS}^{10}$. We are in addition considering different variations of the BOBAX method as described in Section \ref{app:bobax_variants}: We consider $\textbf{EIBAX}^{20}$, $\textbf{EIBAX}^{50}$, $\textbf{EIBAX}^{10}$, as well as $\textbf{BOBAX}^{0.5}_\text{prob}$. Also, we have run BOBAX for different degrees of random interleaving $k \in \{2, 5, 10\}$, denoted by $\textbf{BOBAX}^2, \textbf{BOBAX}^5, \textbf{BOBAX}^{10}$.

Note that all (BAX) variants optimize for a PD for one variable only; we have chosen the first variable as default. To support our claims in Section \ref{par:multiple_variables} that our method can be easily applied to jointly compute the PDP for multiple variables, we are also comparing to one variant which computes the PDP for \emph{all} variables, denoted by $\textbf{BAX}_\textrm{all}$ and compare it to $\textbf{BAX}$. 

\paragraph{Technical details}

All experiments only require CPUs (and no GPUs) and were computed on a Linux cluster (see Table~\ref{tab:cluster}).  

\begin{table}[ht]
\caption{Description of the infrastructure used for the experiments in this paper. }
\label{tab:cluster}
\centering
    \begin{tabular}{ll}
        \toprule
         \multicolumn{2}{c}{Computing Infrastructure} \\ \midrule
         Type & Linux CPU Cluster \\ 
         Architecture & 28-way Haswell-EP nodes \\
         Cores per Node & 1 \\
         Memory limit (per core) & 2.2 GB \\ \bottomrule
    \end{tabular}
\end{table}

\paragraph{Implementation details} Our implementation of BOBAX is based on the implementation provided by \citep{neiswanger2021bax}\footnote{https://github.com/willieneis/bayesian-algorithm-execution}, which in turn is based on the GPflow \citep{matthews17gpflow} implementation for Gaussian processes. 

Note that we are not optimizing the hyperparameters of the GP (lengthscale, kernel variance, and nugget effect) during BOBAX to eliminate one source of variance between methods. Instead, similarly to \citep{neiswanger2021bax}, we are setting those parameters to sensible default values. These are determined by the following heuristic executed prior to all experiments: For every objective function, we perform maximum likelihood optimization of these GP hyperparameters based on $200$ randomly sampled points, and choose the configuration with the highest likelihood. This configuration is fixed across all replications and methods. While this heuristic does not impact the expressiveness of our statements since all methods are based on the same kernel hyperparameters, we emphasize that choosing appropriate hyperparameters is crucial for the performance of our method; therefore, a stable implementation (as done in established BO libraries) is regarded a necessary requirement for practical usage.    

\subsection{Additional Results}
\label{app:benchmark_results}

\begin{figure}[htb]
    \centering
    \includegraphics[width=0.5\textwidth]{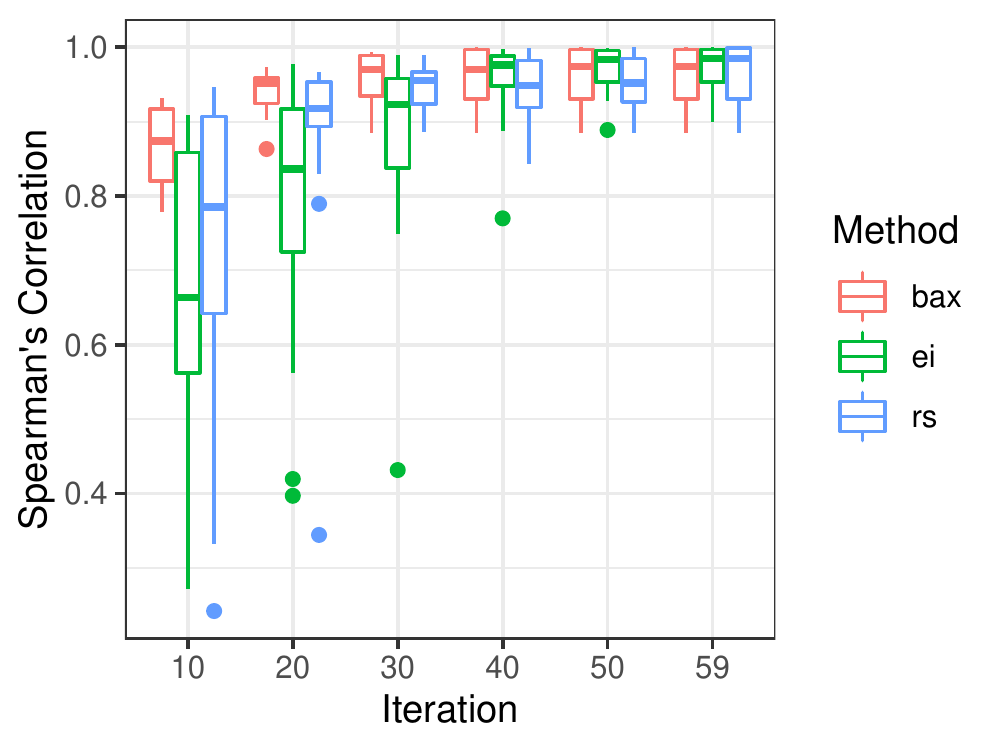}
    \caption{The figure shows Spearman's rank correlation of the estimated PDP vs. the iterations performed for the Branin function. It demonstrates that looking at PDPs computed on data from BO with expected improvement can be even wrong in terms of correlation (as compared to BAX and RS), which matters a lot in the context of optimization. }
    \label{fig:ranks}
\end{figure}

First of all, to provide some evidence for our claim that that BO with EI can return inaccurate PDPs not only in absolute terms but also when considering ranks, we have computed Spearman's rank correlation of the respective PD estimate with the ground truth objective (see Figure \ref{fig:ranks}). 

To evaluate many different algorithms based on two criteria (1) error in PDP estimate $\textrm{d}_{\textrm{dL}_1}$ and (2) optimization regret in a compressed way, we are looking at the ranks of different methods with regards to both metrics, resulting in two ranks $\textrm{rank}_{\textrm{d}_{\textrm{dL}1}}$, $\textrm{rank}_\textrm{regret}$. For the sake of evaluation we assume that interpretability and optimization efficiency are of equal importance and therefore assign each method a combined rank of $\frac{1}{2}\cdot \textrm{rank}_{\textrm{d}_{\textrm{dL}1}} + \frac{1}{2}\cdot \textrm{rank}_\textrm{regret}$. We average the combined ranks of every method across replications and problem instances. Table \ref{tab:combined_ranks} shows the combined ranks for our proposed methods BAX and BOBAX (introduced in Section \ref{alg:bobax}) as well as all baselines.

% As addition to the results presented in Section \ref{sec:benchmark}, we present progress diagrams for the methods compared in Figure \ref{fig:part1_progress}, as well as box plots for the remaining objective functions in Figure \ref{fig:part2_boxplot}. 

\begin{figure*}
    \centering
    \includegraphics[width=0.8\textwidth]{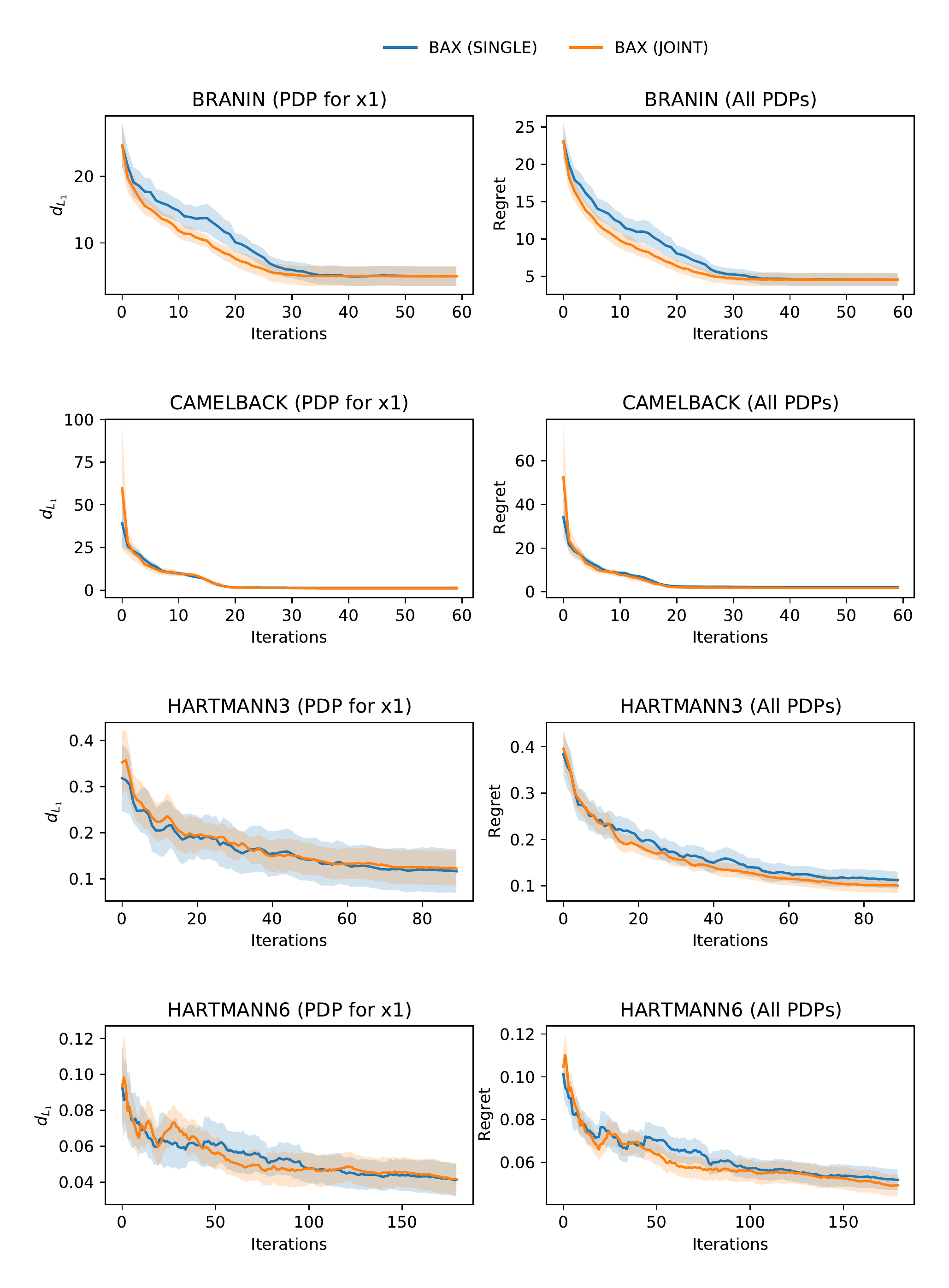}
    \caption{The performance of BOBAX with $\textrm{EIG}_\textrm{PDP}$ computed with regards to the first variable only (blue) vs. the performance of BOBAX when $\textrm{EIG}_\textrm{PDP}$ is computed for the joint execution paths of PD estimates with regards to \emph{all} variables (orange).  \textbf{Left}: Error of the PD estimate for the \emph{first} variable (measured via $\textrm{d}_{\textrm{L}_1}$). \textbf{Right}: Error of the PD estimate for the \emph{all} variables (measured via $\textrm{d}_{\textrm{L}_1}$). We observe that the joint computation delivers more accurate PDs over \emph{all} variables. However, we also observe that the difference is not dramatically big. }
    \label{fig:progress_different_variables}
\end{figure*}

Figure \ref{fig:progress_different_variables} compares the $\textrm{EIG}_\textrm{PDP}$ computed w.r.t. the PD of a single variable vs. jointly for the PDs of all variables. We observe that there is no drop in performance; in particular, we observe that the joint computation performs comparably to the computation for a single variable when evaluated on a single variable; and the joint computation performs better, if the accuracy of \emph{all} PDPs is considered. 

\begin{table}[]
    \caption{The table shows the combined ranks $\frac{1}{2}\cdot \textrm{rank}_{\textrm{d}_{\textrm{dL}1}} + \frac{1}{2}\cdot \textrm{rank}_\textrm{regret}$ of different methods introduced in Section \ref{sec:methods} as well as additional baselines introduced in Appendix \ref{app:benchmark_details}. Results are averaged across $20$ replications and across all problems. We observe that $\text{BOBAX}^2$ is best in terms of the combined rank. }\vspace*{0.1cm}
    \label{tab:combined_ranks}
    \centering
    \begin{tabular}{lrrrr}
    \toprule
     & \multicolumn{4}{c}{Combined ranks after} \\ \cmidrule(l){2-5} 
     & 25\% & 50\% & 75\% & 100\% \\ 
     & \multicolumn{4}{c}{Max. iterations spent} \\ \midrule
    BOBAX$^2$        &            \textbf{6.30} &           \textbf{5.24} &            \textbf{4.88} &         \textbf{4.88} \\
    BOBAX$^5$        &            6.36 &           5.91 &            5.23 &         5.08 \\
    BOBAX$^{10}$       &            6.51 &           6.10 &            5.46 &         5.14 \\
    BO-RS$^2$         &            7.65 &           7.02 &            5.88 &         5.49 \\
    BOBAX$_\textrm{prob}^{0.5}$ &            6.96 &           6.39 &            5.92 &         5.72 \\
    BO-RS$^5$         &            7.24 &           6.60 &            5.92 &         5.73 \\
    BO-RS$^{10}$        &            7.37 &           6.64 &            6.18 &         5.78 \\
    EIBAX$^{100}$             &            6.71 &           6.40 &            6.00 &         5.94 \\
    BAX                     &            6.77 &           6.95 &            6.32 &         6.18 \\
    EIBAX$^{20}$                &            6.91 &           6.79 &            6.10 &         6.20 \\
    LCB$^5$                   &            8.93 &           6.65 &            6.09 &         6.22 \\
    EI                      &            7.53 &           7.44 &            6.67 &         6.28 \\
    EIBAX$^{50}$                &            6.83 &           6.45 &            6.21 &         6.49 \\
    RS                      &            8.82 &           9.19 &            7.74 &         6.92 \\
    PVAR                    &            9.80 &           8.01 &            7.18 &         7.00 \\
    LCB$^2$                   &            8.94 &           7.30 &            7.30 &         7.54 \\
    LCB$^1$                   &            8.36 &           7.99 &            8.20 &         8.10 \\
    \bottomrule
    \end{tabular}
\end{table}

% \begin{figure*}[htb]
%    \centering
%    \includegraphics[width=0.8\textwidth]{figures/part2/progress_main_variants.pdf}
%    \caption{\textbf{Left}: Progress of error in PDP as measured by $\textrm{dL}_1$ vs. the number of iterations. \textbf{Right}: Progress of optimization regret vs. number of iterations. In both figures, the $95\%$ confidence intervals around the mean are shown.}
%    \label{fig:part1_progress}
%\end{figure*}

\begin{figure*}[htb]
    \centering
    \includegraphics[width=0.6\textwidth]{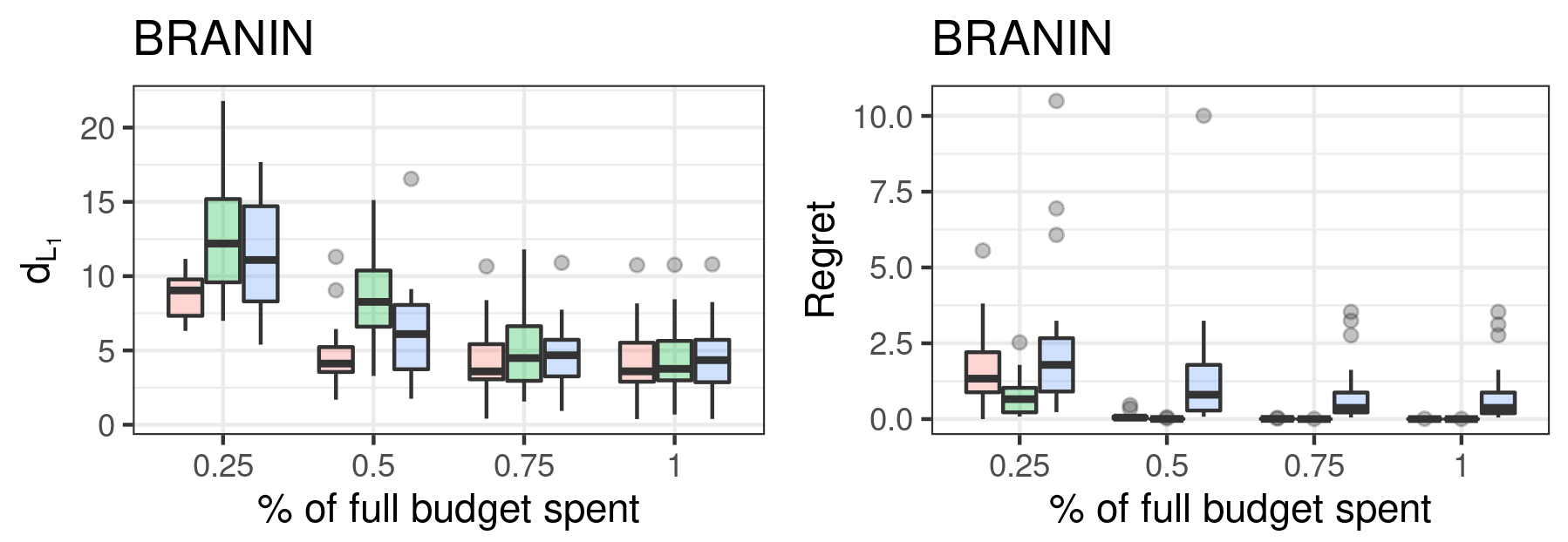} \\
    \includegraphics[width=0.6\textwidth]{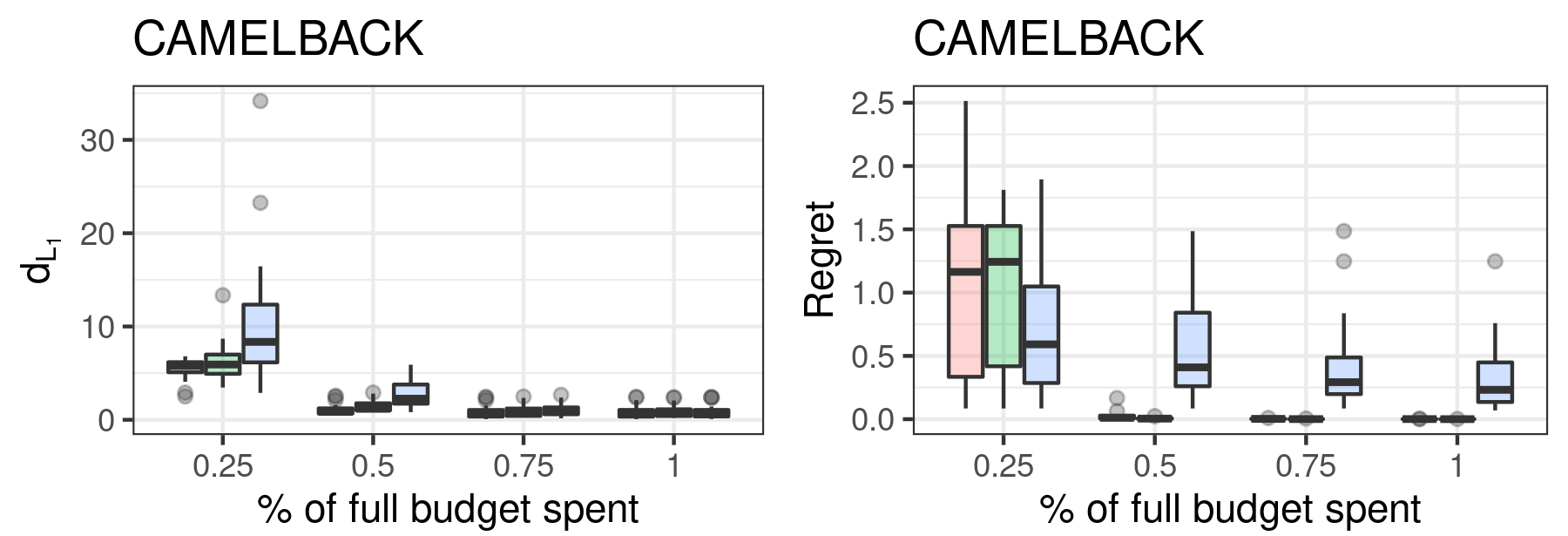} \\
    \includegraphics[width=0.6\textwidth]{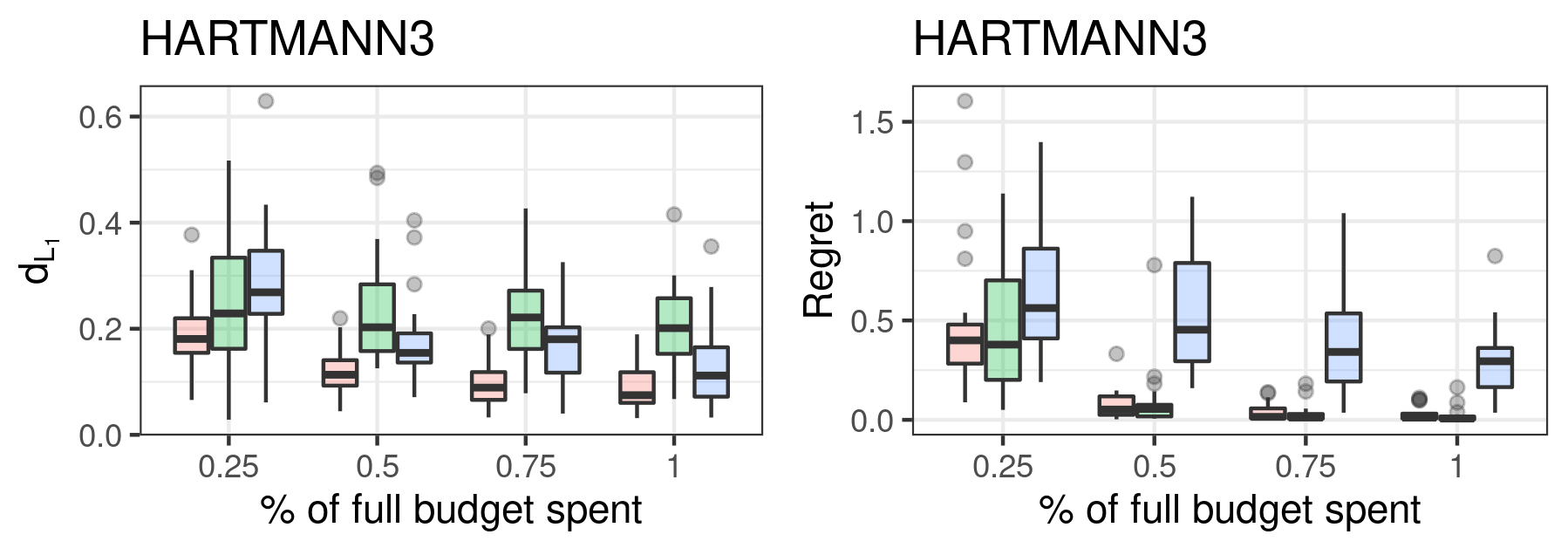} \\
    \includegraphics[width=0.6\textwidth]{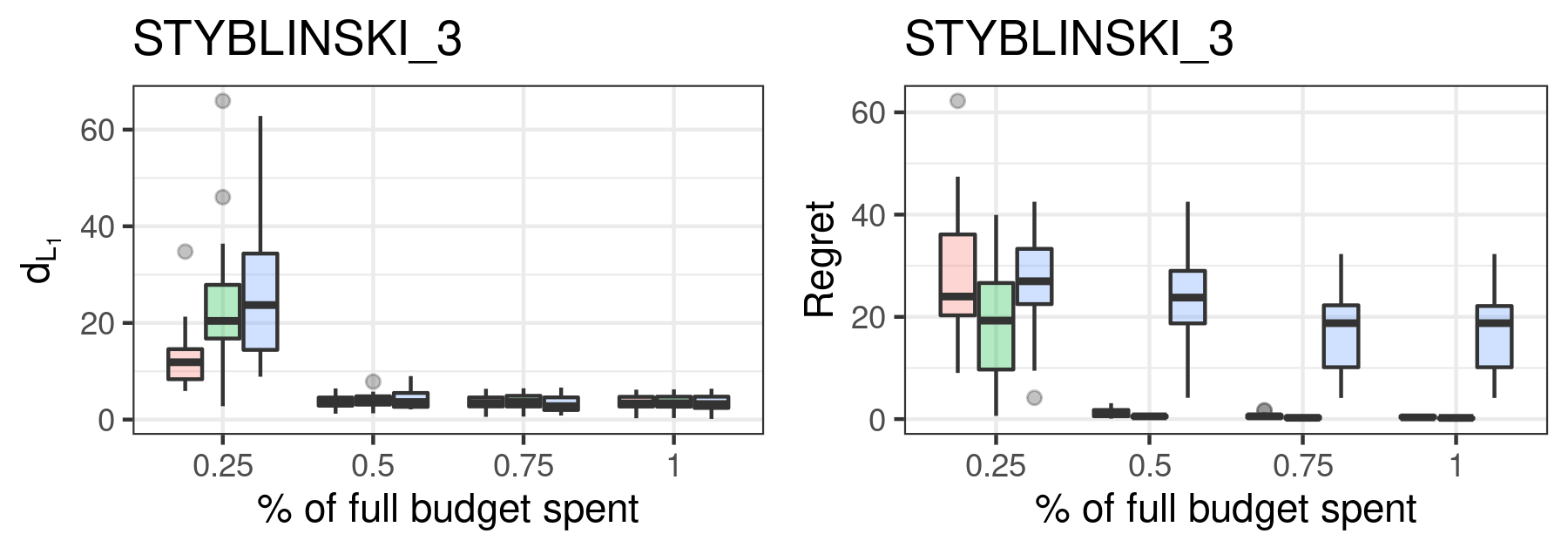} \\
    \includegraphics[width=0.6\textwidth]{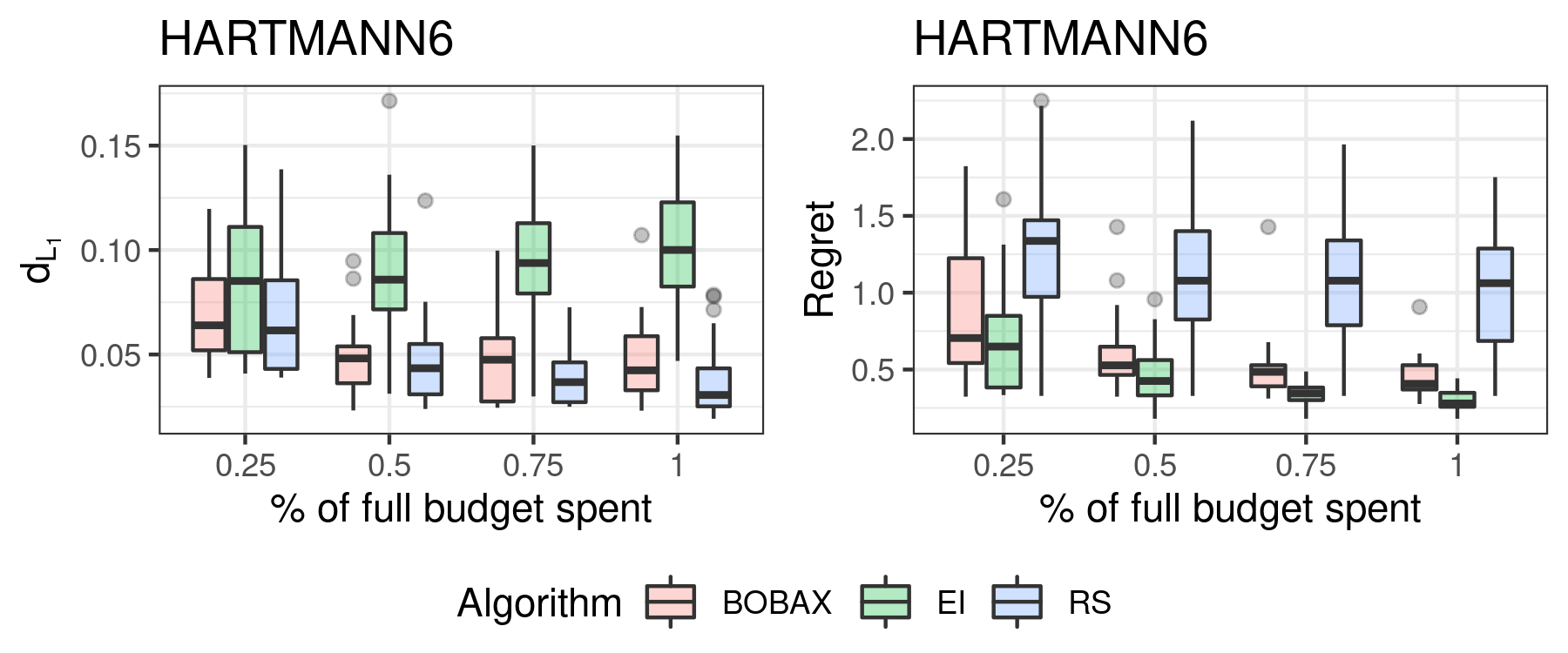}
    \caption{Error of PD estimates measured via measured by $\textrm{d}_{L_1}$ (left) and optimization regret (right) for the different synthetic objectives. While RS is clearly outperformed in terms of optimization efficiency, BOBAX and BO with EI perform comparable on this problem instance. }
    \label{fig:part2_boxplot}
\end{figure*}

\section{Practical HPO Application}
\label{app:practical_HPO}

\subsection{Additional Details}
\label{app:practical_HPO_Details}

\begin{table}[hb]
\centering
    \caption{Hyperparameter space of the LCBench \citep{Zimmer2020AutoPyTorchTM} benchmark suite within YAHPO gym \citep{pfisterer21yahpo}; \emph{batch size} and \emph{maximum number of layers} have been set to defaults $512$ and $5$, respectively. } \vspace*{0.2cm}
    \label{tab:searchspace}
    \begin{tabular}{cccc}
        \toprule
         Name & Range & log & type \\ \midrule
         Max. number of units & $[64, 512]$ & yes & int \\
         Learning rate (SGD) & $[1\textrm{e}^{-4}, 1\textrm{e}^{-1}]$ & yes & float \\
         Weight decay & $[1\textrm{e}^{-5}, 1\textrm{e}^{-1}]$ & no & float \\
         Momentum & $[0.1, 0.99]$ & no & float \\
         Max. dropout rate & $[0.0, 1.0]$ & no & float \\ 
         \bottomrule
    \end{tabular}
\end{table}

\begin{table*}[htb]
\centering
    \caption{Datasets accessed via the \emph{lcbench} suite of YAHPO gym \citep{pfisterer21yahpo}; the underlying data for the surrogate benchmark was made available by \citep{Zimmer2020AutoPyTorchTM}. }
    \label{tab:datasets}
    \begin{tabular}{ccccc}
        \toprule
         ID & Name & Usecase & $n$ & $d$ \\ \midrule
         3945 & KDDCup09\_appetency & Prediciton of customer behavior & 50000 & 231 \\
        34539 & drug-directory & Drug classification & 120215 & 21  \\
        7593 & covertype & Forest cover type & 581012 & 55 \\
        126025 & adult & Salary prediction & 48842 & 15 \\
        126026 & nomao & Active-learning in real-world & 34465 & 119 \\
        126029 & bank-marketing & Bank direct marketing & 4521 & 17 \\
        146212 & shuttle &  & 58000 & 10 \\
        167104 & Australian & Credit approval & 690 & 15 \\
        167149 & kr-vs-kp & Chess game & 3196 & 37 \\
        167152 & mfeat-factors & Handwritten numerals & 2000 & 217 \\
        167161 & credit-g & Credit risk prediciton & 1000 & 21 \\
        167168 & vehicle & Classification of vehicles & 846 & 22 \\
        167185 & cnae-9 & Classification of free text & 1080 & 857 \\
        167200 & higgs & Higgs boson detection & 98050 & 29 \\
        189908 & Fashion-MNIST & Classification of Zalando's article images & 70000 & 785 \\
         \bottomrule
    \end{tabular}
\end{table*}

As practical HPO application we have chosen the use case of tuning hyperparameters of a neural network (as shown in Table \ref{tab:searchspace}) on the different classification tasks (listed in Table \ref{tab:datasets}) with regards to \emph{Balanced accuracy} as performance measures. In BAX / a-BOBAX, we are computing the $\textrm{EIG}_\textrm{PDP}$ jointly for the PDPs of all hyperparameters listed in Table \ref{tab:searchspace}. Each run is replicated $10$ times. Otherwise, all other settings correspond to the settings in Sections \ref{sec:benchmark} and Appendix \ref{app:benchmark}. Note that the benchmark provided via Yahpo Gym \citep{pfisterer21yahpo} is a surrogate benchmark, which not only supports efficient execution of a benchmark, but also gives access to a (reasonably cheap-to-evaluate) empirical performance model as ground truth objective; allowing us to compute the ground-truth PDP (and thus, any measure of error of the PDP) based on this empirical performance model. 

\subsection{Additional Results}
\label{app:practical_HPO_additional_results}

\begin{figure*}
    \centering
    \includegraphics{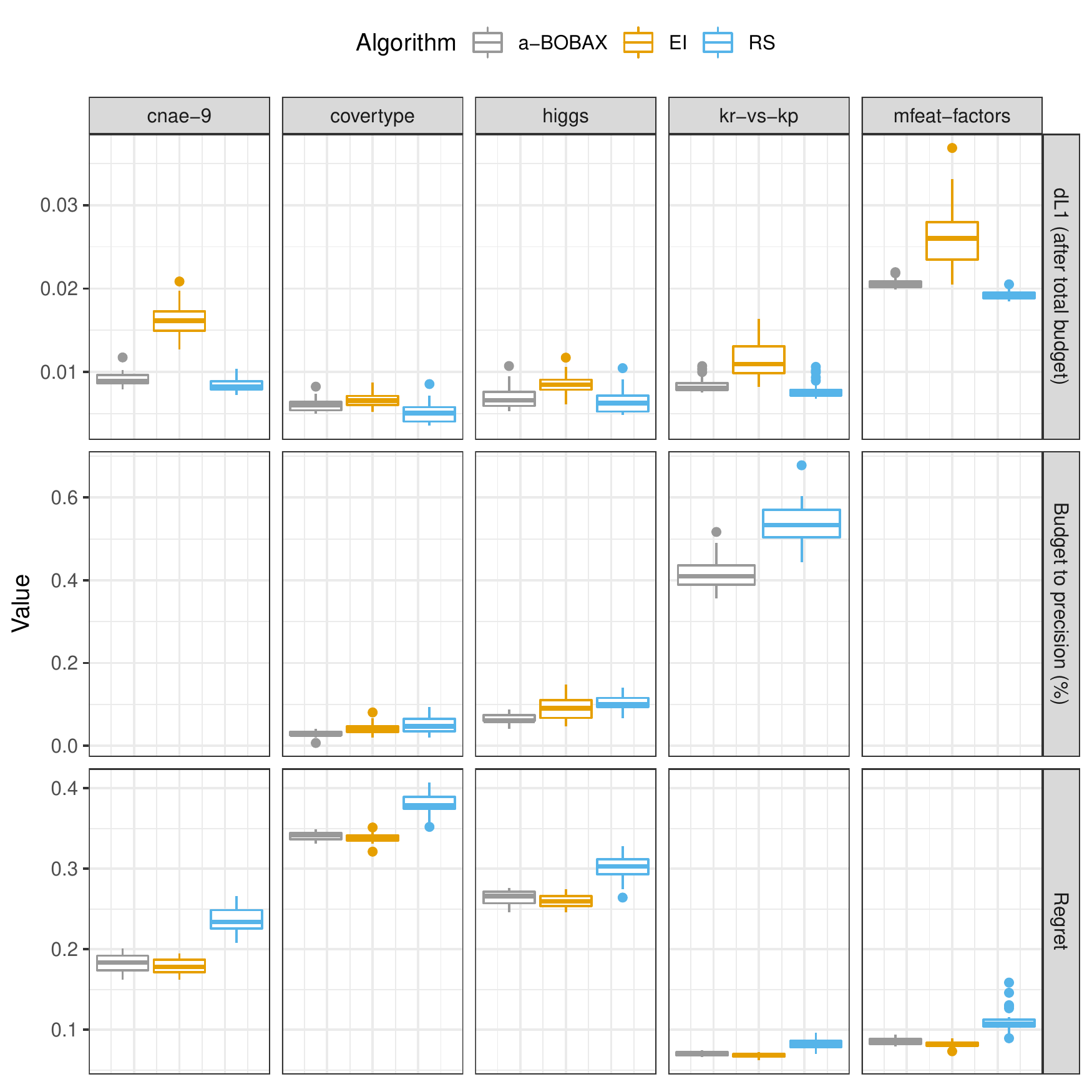}
    \caption{The figure compares error of the PDP estimate after the full budget spent (in terms of $\textrm{dL}_1$; shown in the first row), the percentage of iterations needed to reach the desired level of confidence (middle row), as well as the final regret (last row) for the different methods a-BOBAX, EI, and RS on the different datasets (columns) that we tuned for. In most cases, a-BOBAX has a final error in PDP comparable to RS, but clearly better than with EI, and reaches the desired level of confidence faster then the two other methods. In terms of optimization performance, a-BOBAX and EI perform comparably, and both clearly outperform RS. }
    \label{fig:detailed_results_lcbench1}
\end{figure*}

\begin{figure*}
    \centering
    \includegraphics{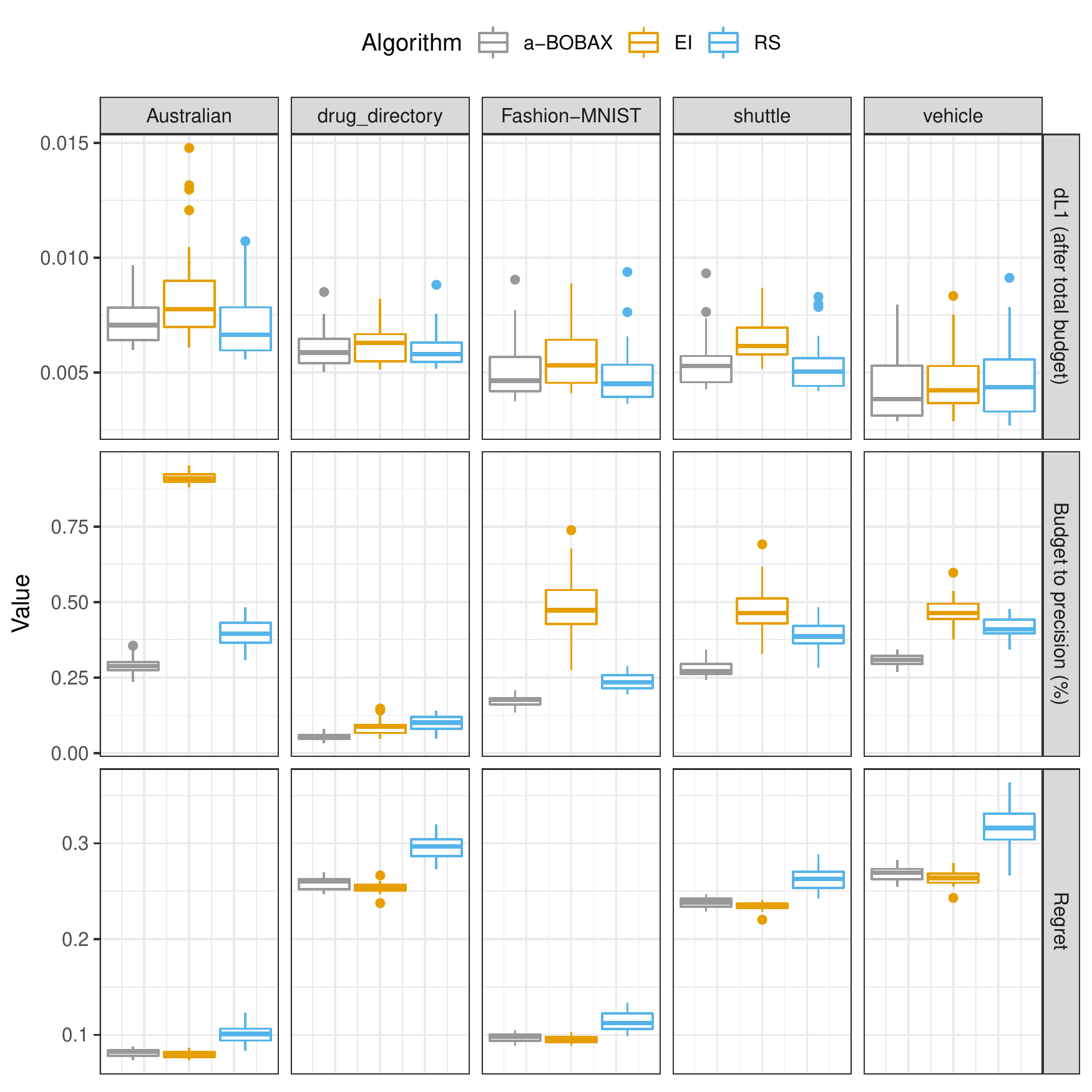}
    \caption{The figure compares error of the PDP estimate after the full budget spent (in terms of $\textrm{dL}_1$; shown in the first row), the percentage of iterations needed to reach the desired level of confidence (middle row), as well as the final regret (last row) for the different methods a-BOBAX, EI, and RS on the different datasets (columns) that we tuned for. In most cases, a-BOBAX has a final error in PDP comparable to RS, but clearly better than with EI, and reaches the desired level of confidence faster then the two other methods. In terms of optimization performance, a-BOBAX and EI perform comparably, and both clearly outperform RS. }
    \label{fig:detailed_results_lcbench2}
\end{figure*}

\begin{figure*}
    \centering
    \includegraphics{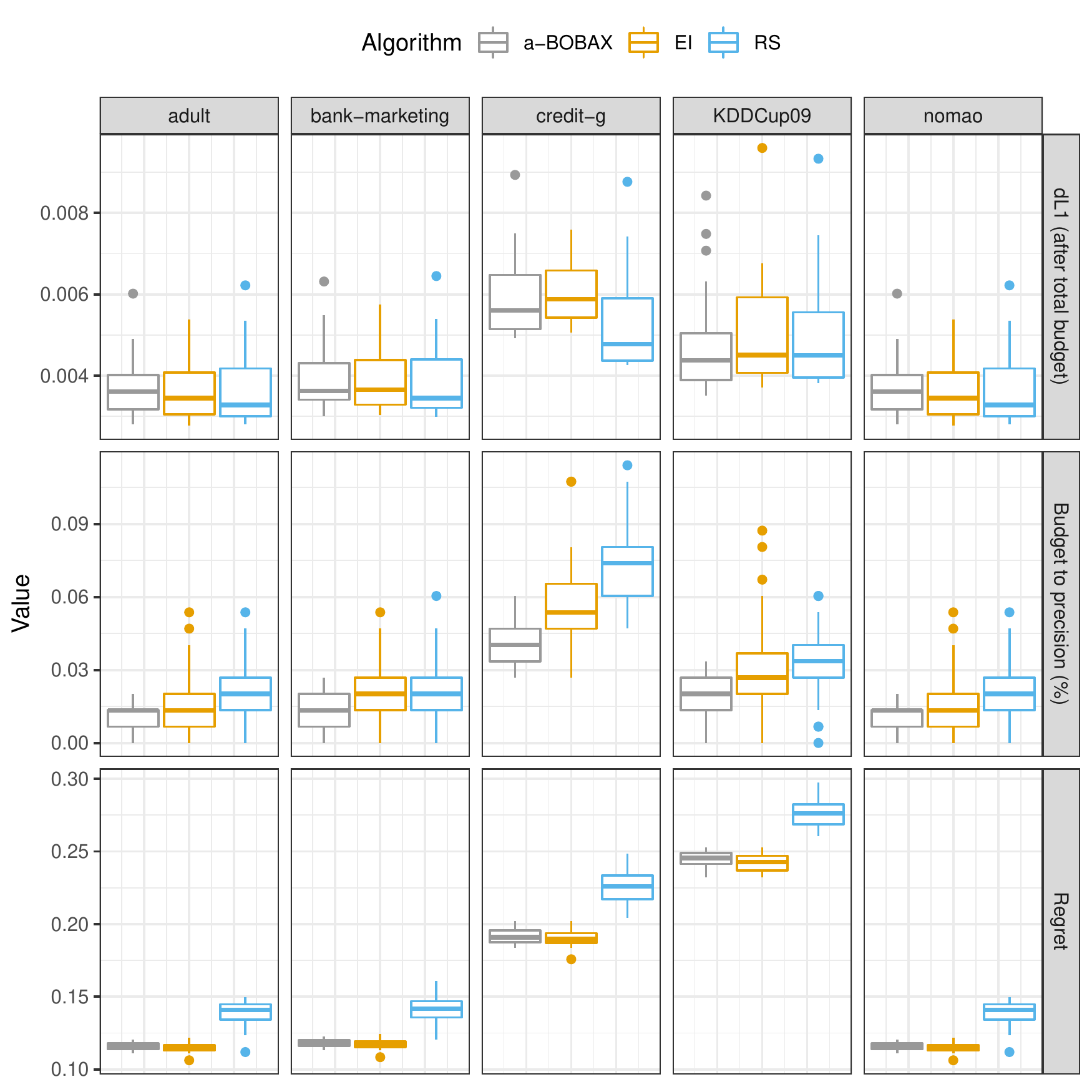}
    \caption{The figure compares error of the PDP estimate after the full budget spent (in terms of $\textrm{dL}_1$; shown in the first row), the percentage of iterations needed to reach the desired level of confidence (middle row), as well as the final regret (last row) for the different methods a-BOBAX, EI, and RS on the different datasets (columns) that we tuned for. In most cases, a-BOBAX has a final error in PDP comparable to RS, but clearly better than with EI, and reaches the desired level of confidence faster then the two other methods. In terms of optimization performance, a-BOBAX and EI perform comparably, and both clearly outperform RS. }
    \label{fig:detailed_results_lcbench3}
\end{figure*}

Figures \ref{fig:detailed_results_lcbench1}, \ref{fig:detailed_results_lcbench2}, \ref{fig:detailed_results_lcbench3} shows a more granular representation of results for the HPO usecase.

\section{Code and Implementation}
\label{app:code}

All code and data needed to reproduce the benchmark will be made publicly available via a Github repository after completion of the review process. During review phase, all code is uploaded as a supplementary material, or can alternatively be downloaded from \url{https://figshare.com/s/d6ef1b8f4c9c1e844229}. Please refer to the README.md file for further information about how to use the code to reproduce results.  

Note that our implementation is based on the implementation provided by \cite{neiswanger2021bax}\footnote{https://github.com/willieneis/bayesian-algorithm-execution}. 

Raw and processed results can be downloaded from \url{https://figshare.com/s/4573a2546f1d8a535c12}.

\end{document}